\documentclass[pdflatex,sn-apa]{sn-jnl}

\jyear{2022}%

\usepackage[utf8]{inputenc} 
\usepackage[T1]{fontenc}    
\usepackage{hyperref}       
\usepackage{url}            
\usepackage{booktabs}       
\usepackage{amsfonts}       
\usepackage{nicefrac}       
\usepackage{microtype}      
\usepackage{color}         

\usepackage{mathtools}
\usepackage{algorithm}
\usepackage{algpseudocode}
\usepackage{amsmath,amsfonts,amssymb,amsthm,mathrsfs}
\usepackage{graphicx}
\usepackage{caption}
\usepackage{subcaption}
\usepackage{multirow}

\newcommand{\red}{\textcolor{red}}

\newcommand{\argmin}{\operatornamewithlimits{argmin}}
\newcommand{\argmax}{\operatornamewithlimits{argmax}}

\newcommand{\BEA}{\begin{eqnarray}}
\newcommand{\EEA}{\end{eqnarray}}
\newcommand{\BEQ}{\begin{equation}}
\newcommand{\EEQ}{\end{equation}}
\newcommand{\BIT}{\begin{itemize}}
\newcommand{\EIT}{\end{itemize}}

\newcommand{\MC}{\mathcal}
\newcommand{\MB}{\mathbb}

\newcommand{\MBE}{\mathbb{E}}
\newcommand{\MBP}{\mathbb{P}}

\newcommand{\NN}{\nonumber}
\newcommand{\TD}{\widetilde}
\newcommand{\WH}{\widehat}

\newcommand{\tr}{\mathrm{tr}}

\theoremstyle{thmstyleone}%
\newtheorem{theorem}{Theorem}
\theoremstyle{thmstyletwo}%
\newtheorem{remark}{Remark}%

\theoremstyle{thmstylethree}%
\newtheorem{definition}{Definition}%

\newtheorem{lemma}[theorem]{Lemma}

\title[Semi-Parametric Contextual Bandits with Graph-Laplacian Regularization]{Semi-Parametric Contextual Bandits with Graph-Laplacian Regularization}

\author[1]{\fnm{Young-Geun} \sur{Choi}}\email{ygchoi@sm.ac.kr}

\author[2]{\fnm{Gi-Soo} \sur{Kim}}\email{gisookim@unist.ac.kr}

\author[3]{\fnm{Seunghoon} \sur{Paik}}\email{shpaik@berkeley.edu}

\author*[4]{\fnm{Myunghee Cho} \sur{Paik}}\email{myungheechopaik@snu.ac.kr}

\affil[1]{\orgdiv{Department of Statistics}, \orgname{Sookmyung Women's University}, \orgaddress{\state{Seoul 04310}, \country{Rep. of Korea}}}

\affil[2]{\orgdiv{Department of Industrial Engineering and Graduate School of Artificial Intelligence}, \orgname{UNIST}, \orgaddress{\state{Ulsan 44919}, \country{Rep. of Korea}}}

\affil[3]{\orgdiv{Department of Statistics}, \orgname{University of California Berkeley}, \orgaddress{\state{CA 94720}, \country{USA}}}

\affil*[4]{\orgdiv{Department of Statistics}, \orgname{Seoul National University}, \orgaddress{\state{Seoul 08826}, \country{Rep. of Korea}}}

\abstract{Non-stationarity is ubiquitous in human behavior and addressing it in the contextual bandits is challenging.  Several works have addressed the problem by investigating semi-parametric contextual bandits and warned that ignoring non-stationarity could harm performances.  Another prevalent human behavior is social interaction which has become available in a form of a social network or graph structure. As a result, graph-based contextual bandits  have received much attention.  In this paper, we propose {\sf SemiGraphTS}, a novel contextual Thompson-sampling algorithm for a graph-based semi-parametric reward model.  Our algorithm is the first to be proposed in this setting. We derive an upper bound of the cumulative regret that can be expressed as a multiple of a factor depending on the graph structure and the order for the semi-parametric model without a graph.  We evaluate the proposed and existing algorithms via simulation and real data example.}


\begin{document}

\maketitle

\section{Introduction}
\label{sec:intro}

In contextual multi-armed bandits (MAB),  a learning agent sequentially chooses actions while balancing to maximize the reward (exploitation) and  to learn the reward mechanism as a function of contexts with higher precision (exploration). Algorithms for contextual MAB problems have demonstrated their usefulness in many applications including recommendations of news articles, advertisements, or behavioral interventions \citep{Li2010a, Tang2013,Tewari2017a}. Thompson sampling (TS)-based algorithms randomly choose an action from repeatedly updated posterior, and have been widely used among other bandit algorithms \citep{Scott2010,Kaufmann2012,Agrawal2013}.

The semi-parametric contextual bandit \citep{Greenewald2017,Krishnamurthy2018,Kim2019} models  the mean of the reward by a 
linear function of the contexts and 
a time-varying intercept. 
The algorithms for semi-parametric models 
allow the reward distribution to change over time in a non-stationary manner. For example, behavior may change over time depending on the user's  circumstances or preference for a shopping item may change according to a time trend.  These may not be captured in the context vectors.
In the single-user setting, the semi-parametric bandits have demonstrated success in accommodating non-stationarity in mobile health and product recommendation  \citep{Greenewald2017,Kim2019,Peng2019,Liao2019}.

In many real-life settings, there are multiple users and the relationships among the users in a social network are often available as side information.
Such graph information has been utilized in recommendation \citep{Li2010a,Delporte2013,Rao2015}. 
Several graph-based contextual MAB algorithms have been proposed to take the graph information into account 
under the ordinary linear reward assumption \citep{Casa-bianchi2013,Gentile2014,Vaswani2017,Li2019,Yang2020}.  The aforementioned graph-based methods have shown to take advantages of a graph structure and perform well, but  may be restrictive in real-life settings when the rewards tend to change over time. 

Our goal is to construct a semi-parametric bandit algorithm that accommodates multiple users equipped with a network, with practically feasible computational cost. To the best of our knowledge, our algorithm is the first algorithm proposed in this setting. The main contributions of the work presented in this paper are as follows.
\BIT
\item We propose {\sf SemiGraphTS} (semi-parametric-graph-Thompson-sampling), a novel TS algorithm for a setting in which each user's reward follows the semi-parametric model and user-specific parameters are regularized by the given graph. 

\item We derive an upper bound of the cumulative regret for {\sf SemiGraphTS}, which be expressed as a multiple of a factor depending on the graph structure and the bound from  the semi-parametric model without a graph.

\item We propose a novel scalable estimator for the user-specific parameter that incorporates the estimators from the neighbors defined by the graph structure while conditioning out time-dependent coefficients. This plays a crucial role in building the {\sf SemiGraphTS} algorithm. 
We establish a high-probability upper bound for its estimation error. 
\EIT

\section{Model and problem setting}
\label{sec:prelim}

We study the semi-parametric contextual bandit problem for multiple users equipped with a user network. 
Suppose that there are $n$ users, say $j \in V = \{ 1, \ldots, n \}$. 
For each time step $t= 1, \ldots, T$, the learning agent is instructed which user to serve, say $j_t$. 
The agent is supposed to recommend an item or pull an arm for the target user based on the previous action history and the contexts describing the items. Suppose that there are $N$ candidate  arms, say $i=1, \ldots, N$, and that a context vector $b_i(t) \in \MB{R}^d$ represents the feature of  the $i$-th item at time $t$. 
We denote by $a(t)$ the selected arm  to recommend to the target user.
We let $r_{i,j}(t)$ be the reward for arm $i$, user $j$ at time $t$.
Upon the action, the user returns a user-specific reward for the chosen arm, say $r_{a(t), j_t} (t)$.
The information given to the learner at time $t$ is formally described as filtration $\MC{F}_{t-1} = \{ j_t, \{ b_i(t)\}_{i=1}^N \} \bigcup \left( \bigcup_{\tau=1}^{t-1} \{ j_{\tau}, \{ b_i(\tau)\}_{i=1}^N, a({\tau}), r_{a(\tau), j_{\tau}} (\tau) \} \right)$.

The multiple-user semi-parametric reward model is described as below:
\BEQ\label{eqn:rewardSemi}
r_{i,j}(t) = \nu_j(t) + b_i(t)^T \mu_j + \eta_{i,j}(t),
\EEQ
for $i=1,\ldots,N$, $j=1,\ldots,n$, and $t=1,\ldots,T$.
Here, $\mu_j \in \MB{R}^d$ denotes the  unknown user-specific parameter that represents the preference of the $j$-th user for a given context.
The intercept $\nu_j(t)$ indicates the baseline reward for user $j$ at time $t$. We do not impose any parametric assumption on the functional form of $\nu_j(t)$;  we allow the baseline to arbitrarily change over time and users, whatever gradually and abruptly.   When $\nu_j(t) = 0$ for all $j$, \eqref{eqn:rewardSemi} is reduced to the standard linear reward model. 
Without loss of generality, we assume a uniform boundedness of the contexts and true parameters, i.e., $ \lvert\nu_j (t)\rvert \leq 1 $, $\| b_i(t) \| \leq 1$ and $\| \mu_j \| \leq 1$ for all $i$, $j$ and $t$, where $\|\cdot\|$ denotes the vector $\ell_2$ norm. This assumption can be satisfied by rescaling the data.
We assume that the random error $\eta_{i,j}(t)$ satisfies $\MBE ( \eta_{i,j}(t) \vert \MC{F}_{t-1} ) = 0$. If $n=1$, \eqref{eqn:rewardSemi} coincides with  the single-user semi-parametric bandit problem.

The optimal arm $a^*(t)$ is defined as the arm that maximizes the expected reward for the $j_t$-th user given the history, that is, $a^*(t) = \argmax_i \MBE( r_{i,j_t}(t) \vert \MC{F}_{t-1}) = \argmax_i \{ \nu_{j_t}(t) + b_i(t)^T \mu_{j_t} \} = \argmax_i \{ b_i(t)^T \mu_{j_t} \}$.
Although $a^*(t)$ may be different across users, in each round $t$, only one user enters, and we omit the subscript.
Regret at time $t$ is defined by the difference between the expected rewards from the optimal arm and the chosen arm,
\BEA
regret(t) &=&  \MBE( r_{a^*(t),j_t}(t) \vert \MC{F}_{t-1}) - \MBE( r_{a(t),j_t}(t) \vert \MC{F}_{t-1}) \NN \\ 
&=&
b_{a^*(t)}(t)^T \mu_{j_t}  -  b_{a(t)}(t)^T \mu_{j_t}.\NN
\EEA
The goal of the agent is to minimize the cumulative regret, 
$R(T) = \sum_{t=1}^{T}   regret(t)$.

In graph-based bandit settings, the user network is given a priori as the side information. %
Without any information on the user network, the problem reduces to learning $n$ independent instances.
Let $\MC{G} = (V,E)$ be an undirected simple graph, where a node $j \in V = \{1, \ldots, n\}$ corresponds to a user and an edge $\{j,k \} \in E$ represents the link between users. 
There are several ways to uniquely represent $\MC{G}$ as a Laplacian matrix $L=(l_{jk}) \in \MB{R}^{n \times n}$. We employ the random-walk normalized Laplacian defined by 
\BEQ
l_{jj}=1,~ ~~~ l_{jk} = \begin{cases} -1/\deg(j) & \mbox{if $\{j,k\} \in E$,} \\  0 & \mbox{otherwise,} \end{cases} \label{eqn:random_walk_Laplacian}
\EEQ
for $j,k = 1,\ldots, n$ with $j\neq k$.
In addition, let $\Delta_j = \sum_{k=1}^n l_{jk}\mu_k = \mu_j - \sum_{k:\{j,k\} \in E} \mu_k / \deg(j)$. 
The choice of random-walk normalized Laplacian is particularly useful in the regret analysis and discussed after the proof sketch. 
Our working assumption is that $\|\Delta_j\|$ is small for all $j$, i.e., the edges encode the affinity of user preferences. 
Without loss of generality, we assume that $\MC{G}$ is connected. If not, each connected component of users do not share any information of parameters and it suffices to learn each connected component separately. 

In addition, let   $\|x\|_A = \sqrt{x^T A x}$ for $x \in \MB{R}^d$ and a positive semi-definite $A \in \MB{R}^{d \times d}$. A matrix-valued inequality $A \geq B$ ($A > B$) denotes that $A-B$ is positive semi-definite (positive definite).

\subsection{Related work}
\label{subsec:related}

Since linear contextual MAB problems for single users were investigated \citep{Abbasi-Yadkori2011,Agrawal2013}, there has been a rich line of works on contextual bandits in recent years. For conciseness, we focus on works that consider either the semi-parametric model for single user or the linear model for multiple user equipped with graph.

\paragraph{Semi-parametric contextual MABs for single user.}
The semi-parametric reward model for a single user \citep{Greenewald2017,Krishnamurthy2018,Kim2019} assumes, say
\BEQ\label{eqn:rewardSemiSingle}
r_{i}(t) = \nu(t) + b_i(t)^T \mu + \eta_{i}(t),
\EEQ
which is a special case of our model \eqref{eqn:rewardSemi} with $n=1$. 
\citet{Greenewald2017} first proposed \eqref{eqn:rewardSemiSingle}. 
A novel challenge in the semi-parametric bandit problem is to mitigate the confounding effect from the baseline reward. 
\citet{Greenewald2017} considered a two-stage TS algorithm that  fixes a random base action and contrasts the base and other actions.
\citet{Krishnamurthy2018} proposed another TS algorithm that contrasts every pair of actions repeatedly.
\citet{Kim2019} proposed a single-step TS algorithm and arguably the state-of-the-art in this setting. Specifically, for each time $t$, they estimate $\mu$ in \eqref{eqn:rewardSemiSingle} by
$\WH{\mu} (t) = B (t)^{-1} \sum_{\tau =1}^{t-1}  2 X_{\tau} r_{a(\tau), k}(\tau)$, where $X_{\tau} = b_{a(\tau)}(\tau) - \MBE(b_{a(\tau)}(\tau) \vert \MC{F}_{\tau-1})$ and  $B(t) = \WH{\Sigma}_{t} + \Sigma_{t} + I_d$ where $\WH{\Sigma}_{t} = \sum_{\tau =1}^{t-1} X_{\tau} X_{\tau}^T$, and $\Sigma_{t} = \sum_{\tau =1}^{t-1} \MBE(X_{\tau} X_{\tau}^T \vert \MC{F}_{\tau-1})$.
Compared with \citet{Agrawal2013}, a TS algorithm under the standard linear reward model, the context vector and covariance part were centered by $\MBE(b_{a(\tau)}(\tau) \vert \MC{F}_{\tau-1})$, which is crucial for ruling out the confounding effect of $\nu(t)$. The regret bound derived in \citet{Kim2019} has the same order with that in \citet{Agrawal2013}.

\paragraph{Linear graph-based bandit algorithms for multiple users.}
Algorithms for graph-based linear contextual bandits have been proposed under the following model \citep{Casa-bianchi2013,Gentile2014,Vaswani2017,Li2019,Yang2020,Li2021}:
\BEQ\label{eqn:rewardLin}
r_{i,j}(t) = b_i(t)^T \mu_j + \eta_{i,j}(t),
\EEQ
which coincides with a special case of \eqref{eqn:rewardSemi} when $\nu_j(t) = 0$.
\citet{Gentile2014} proposed an algorithm utilizing the given graph for clustering  users, where those in the same cluster are represented by the same parameter. \citet{Li2019} generalized \citet{Gentile2014}'s algorithm to address non-uniform user frequencies.  \citet{Li2021} proposed another clustering-based algorithm that allows each $\mu_j$ to change abruptly over time. The regret bound proposed in this work depends on the number of abrupt shifts and can be linear in $T$ if the shifts occur proportionally to $T$.
On the other hand, \citet{Casa-bianchi2013} and \citet{Vaswani2017} proposed UCB- and TS-based algorithms with regret bound $\TD{O} (  dn\sqrt{T})$, where  the entire parameters for all users are estimated under regularization by a graph Laplacian. However, this led to scalability issues as a result of solving an equation involving $nd$ by $nd$ matrix. \citet{Yang2020} proposed a local version of the 
\citet{Casa-bianchi2013} with an improved regret bound $\TD{O} ( \Phi d\sqrt{nT})$, where $\Phi \in (0,1)$ depends on $\MC{G}$.
It updates only the parameter associated with the user to serve at each round. Specifically, \citet{Yang2020} first calculates the ordinary least squares estimator $\bar{\mu}_k(t)$ for each user $k$ as if running $n$ bandits independently. Then, $\mu_{j_t}$ is estimated by adjusting $\bar{\mu}_{j_t}$ for $\bar{\mu}_k(t)$ weighted by the Laplacian, particularly
$\WH{\mu}_{j_t} (t) = \bar{\mu}_{j_t} (t) - \lambda C_{j_t} (t)^{-1}   \sum_{k =1}^n   l_{j_t k} \bar{\mu}_{k} (t)$, where $\lambda$ is a tunable parameter and $C_{j_t}$ is the gram matrix of the selected arm features for user $j_t$ up to time $t$.

\section{Proposed Algorithm}
\label{sec:method}

We observe rewards that are correlated with neighbors defined from the given graph structure and yet whose conditional mean changes over time.
Our main challenge is to incorporate the network information in estimating $\mu_j$ while handling the confounding by $\nu_j(t)$.
Our strategy is to handle non-stationarity for each individual by conditioning, while simultaneously accommodating information from neighbors.  The key idea of conditioning is based on that the non-stationarity does not change across the arms, hence centering the context around the mean for the arms does not alter the problem of finding the maximum reward across the arms. This allows us to construct an estimator of $\mu_j$ that is robust to the effect of $\nu_j(t)$ while exploiting the user affinity information via graph.

The proposed {\sf SemiGraphTS} algorithm is described in Algorithm \ref{algo:SemiGraphTS}. Key steps include parameter estimation and Thompson sampling steps.

\begin{algorithm}[h]
\caption{Proposed algorithm ({\sf SemiGraphTS})} \label{algo:SemiGraphTS}
\begin{algorithmic}[1] 
\State Fix $\lambda > 0$. Set $B_j(1) = \lambda l_{jj} I_d$, $y_j(1) = 0_d$  and $v_j = (4R+12) \sqrt{  d  \log \left\{ ({24} T^4  / \delta) (1 + \lambda^{-1} )   \right\}  } + \sqrt{\lambda} (1 + \|\Delta_j\|)$ for $j=1,\ldots,n$.
\For{$t=1,2,\ldots, T$}
    \State{Observe $j_t$.}
    \For{$j=1,2,\ldots, n$}
        \If{$j \neq j_t$} 
            \State Update $B_j (t+1) \gets B_j (t)$, $\bar{\mu}_j (t+1) \gets \bar{\mu}_j (t)$, and $y_j (t+1) \gets y_j(t)$. 
        \Else
            \State $\WH{\mu}_j(t) \gets \bar{\mu}_j (t) - B_j (t)^{-1}   \sum_{k \neq j}  \lambda l_{jk} \bar{\mu}_{k} (t)$.
            \State $\Gamma_j(t) \gets B_j(t)  + \lambda^2 \sum_{k \neq j} l_{j k}^2  B_k(t)^{-1}$
            \State Sample $\TD{\mu}_j(t)$ from $\MC{N}_d( \WH{\mu}_j(t), v_j^2 \Gamma_j(t)^{-1} )$.
            \State Pull arm $a(t) = \argmax_{i} \{ b_i(t)^T \TD{\mu}_j(t) \}$ and get reward $r_{a(t), j} (t)$.
            \State $\pi_i (t) \gets \MBP( a(t) = i 
            \vert \MC{F}_{t-1})$, $i=1,\ldots,N$.
            \State $\bar{b}(t) \!\gets\! \sum_{i=1}^N\! \pi_i (t) b_i (t)$ and $X_t \!\gets\!  b_{a(t)}(t) - \bar{b}(t)$.
            \State Update $B_j (t+1) \!\gets\!  B_j (t) \!+\! X_t  X_t^T \!+\! \sum_{i=1}^N \! \pi_i(t) (b_i(t)-\bar{b}(t))(b_i(t)-\bar{b}(t))^T$, $y_j (t+1) \!\gets\! y_j (t) \!+\! 2 X_t r_{a(t), j} (t)$, and  $\bar{\mu}_j(t+1) \!\gets\! B_j (t+1)^{-1} y_j (t+1)$.
        \EndIf
    \EndFor
\EndFor
\end{algorithmic}
\end{algorithm}

In the  parameter estimation step, we propose a novel estimator $\WH{\mu}_{j_t} (t)$ for the $j_t$-th user, which is constructed as follows. Define $\MC{T}_{j,t} \!=\! \{ \tau : j_{\tau} = j , 1 \leq \tau \leq t \}$, i.e, $\MC{T}_{j,t}$ collects time indices when user $j$ is served up to time $t$.  
We first calculate an unadjusted user-specific estimator $\bar{\mu}_{k}(t)$ ($k=1,\ldots,n$) proposed by
\BEQ \label{eqn:mubarsemi} 
\bar{\mu}_k(t) = B_k (t)^{-1} \!\!\!\!\!  \sum_{\tau \in \MC{T}_{k,t-1}}\!\!\!\!\! 2 X_{\tau} r_{a(\tau), k}(\tau),\\
\EEQ
where
\BEQ\label{eqn:B_jsemi}
B_k(t) = \WH{\Sigma}_{k,t} + \Sigma_{k,t} + \lambda l_{kk} I_d,
\EEQ
$X_{\tau} \!=\! b_{a(\tau)}(\tau) \!-\! \MBE(b_{a(\tau)}(\tau) \vert \MC{F}_{\tau -1})$, $\WH{\Sigma}_{k,t} \!=\! \sum_{\tau \in \MC{T}_{\!k,t-1}}\! X_{\tau} X_{\tau}^T$, and $\Sigma_{k,t} \!=\! \sum_{\tau \in \MC{T}_{k,t-1}}$ $\MBE(X_{\tau} X_{\tau}^T \vert \MC{F}_{\tau-1})$, $k=1,\ldots,n$.
The expectation in $X_\tau$ and $\Sigma_{k,t}$ originates  from the randomness of $a(t)$ given $\MC{F}_{t-1}$. The definition of $\bar{\mu}_k(t)$ coincides with calculating a regularized version of \citet{Kim2019}'s estimator for each user independently. 
Then, the main proposed estimator $\WH{\mu}_{j_t} (t)$ is given by
\BEQ\label{eqn:muhat_propsemi}
\WH{\mu}_{j_t} (t) = \bar{\mu}_{j_t} (t) - 
    \lambda B_{j_t} (t)^{-1}   \sum_{k \neq j_t}   l_{j_t k} \bar{\mu}_{k} (t).
\EEQ
Intuitively, $\WH{\mu}_{j_t} (t)$ adjusts  $\bar{\mu}_{j_t}(t)$ by the neighborhood counterpart according to the graph structure. 
The designation of \eqref{eqn:muhat_propsemi} is motivated from \citet{Yang2020}  and carefully constructed so that  the estimation error can be expressed in terms of three different types of martingales (with respect to $\MC{F}_{t-1}$), $\eta_{j_t,\tau}(\tau)$, $X_{\tau}$, and $D_{\tau}$ as follows:
\BEQ\NN
\WH{\mu}_{j_t} (t) \!-\! \mu_{j_t} \!=\! B_{j_t}(t)^{-1} \Bigg[     c_{j_t} - \lambda \Delta_{j_t} + 
\sum_{k =1}^n  \left\{ M_{j_t k}  \!\!\sum_{\tau \in \MC{T}_{k,t-1}}\!\! \big( X_{\tau} \eta_{j_t,\tau}(\tau) + A_k(\tau) \big) \right\} 
\Bigg],
\EEQ
\noindent
where
$A_k(t)\! =\! \sum_{\tau \in \MC{T}_{k,t\!-\!1}}\!\! D_{\tau} \mu_k\! +\! 
\sum_{\tau \in \MC{T}_{k,t\!-\!1}}\!$ $\! 2X_{\tau} \! \left(\nu_k(\tau)\!+\!\bar{b}(\tau)^T\mu_k\right)$,
$D_{\tau}\! =\! X_{\tau}X_{\tau}^T\!-\! \MBE(X_{\tau}X_{\tau}^T \vert \MC{F}_{\tau-1})$,
$M_{jk} = I_d$ if $j=k$ and  $\lambda l_{jk} B_k(t)^{-1}$ if $j \neq k$, and $c_j$ is a constant term bounded by $\lambda$. Centering induces $X_{\tau}$ which in turn absorbs non-stationary term, $\nu_k(\tau)$. Detailed proof of sketch is provided in the next Section. The tuning parameter $\lambda$ controls the influence of the graph structure. For a larger $\lambda$, \eqref{eqn:muhat_propsemi} indicates that adjacent nodes more profoundly affect on $\WH{\mu}_{j_t}$. Our regret analysis does not make any assumptions based on $\lambda$, except for $\lambda > 0$.

In the Thompson sampling step, we propose to sample 
 $\TD{\mu}_{j_t}(t)$ from $\MC{N}_d( \WH{\mu}_{j_t}(t), v_{j_t}^2 \Gamma_{j_t}(t)^{-1} )$, where 
\BEQ\label{eqn:Gamma}
\Gamma_j(t) = B_j(t)  + \lambda^2 \sum_{k \neq j}  l_{j k}^2 B_k(t)^{-1}.
\EEQ
The choice of $\Gamma_{j_t}(t)$ in the variance part replaces a conventional choice $B_{j_t}(t)$. Since each $B_k(t)$ is positive definite, it holds that  $\Gamma_{j_t}(t)^{-1} < B_{j_t}(t)^{-1}$. This intuitively means that $\Gamma_{j_t}(t)$ contains more information than $B_{j_t}(t)$ by incorporating the neighborhood information. As a result, our proposed sampling searches over narrower region around $\WH{\mu}_{j_t}(t)$ than the sampling with variance $v_{j_t}^2 B_{j_t}(t)^{-1}$. This leads to an improvement of regret up to a factor less than one compared to an algorithm without graph, as we will see in the next Section.
Finally, we select the arm $a(t)$ that satisfies $a(t) = \argmax_{i} \{ b_i(t)^T \TD{\mu}_{j_t}(t) \}$.

It is worth mentioning that the proposed estimator $\WH{\mu}_{j_t}$ and Thompson sampling step are \emph{local}, in a sense that we run the procedure only for user $j_t$ at each time, not for the entire users. The idea of local update appears natural because  we have no updated information about the other nodes at time $t$.

The terms related to the conditional expectation can be calculated as follows. We define $\pi_i (t)$ as the probability of choosing the $i$-th arm at time $t$, that is, $\pi_i (t) = \MBP(a(t) \!=\! i \vert \MC{F}_{t-1})$. This is determined by the posterior distribution of $\TD{\mu}_{j_t}(t)$, which calls for the evaluation of an integral of a multivariate normal density on a polytope.
One may employ well-known approximation algorithms for the integral, for example, \citet{Wilhelm2010} and \citet{Botev2017}. In our experiments on both synthetic and real data, the Monte Carlo approximation performed well. Once $\pi_i(t)$ is obtained, we can calculate $\MBE(b_{a(t)}(t) \vert \MC{F}_{t-1}) = \MBE( \sum_{i=1}^N I(a(t) \!=\! i) b_{i}(t) \vert \MC{F}_{t-1}) = \sum_{i=1}^N \pi_i (t) b_i (t)$. Similarly, $\MBE(X_{t} X_{t}^T \vert \MC{F}_{t-1}) = \sum_{i=1}^N \pi_i (t) ( b_i (t) - \bar{b} (t) ) ( b_i (t) - \bar{b} (t) )^T$, where $\bar{b} (t) = \MBE(b_{a(t)}(t) \vert \MC{F}_{t-1})$.


The computation complexity of the proposed algorithm is $O(d^2 N + d^2 \deg(j_t) + M(d^2 + dN) )$ if we use the Monte Carlo approximation for evaluating $\pi_i(t)$, where $M$ is the number of Monte Carlo samples. Note that the complexity does not depend on $n$; thus, the proposed algorithm is scalable for large graphs, provided that the average degree of nodes is in a moderate range.
To see why, 
 first, $\WH{\mu}_{j_t} (t)$ and $\Gamma_{j_t}(t)$  in \eqref{eqn:muhat_propsemi} requires $O(d^2 \deg(j_t))$ computations given $\bar{\mu}_k(t)$. As for $\bar{\mu}_k(t)$ and $B_k(t)$, note that $B_j(t) = B_j(t-1)$ and $\bar{\mu}_j(t) = \bar{\mu}_j(t-1)$ if $j \neq j_t$. Thus, $\bar{\mu}_k(t)$ and $B_k(t)$ is computed only for $k=j_t$, which requires $O(d^2N)$ operations.
In addition, the Thompson sampling step and the approximation for $\pi_i(t)$ cost $O(M(d^2 + dN))$.
To compare with the fastest algorithms in similar settings, \citet{Kim2019} and \citet{Yang2020} require $O(d^2N + M(d^2 + dN))$ and $O(d^2  \deg(j_t))$ operations, respectively. Although the proposed algorithm has slightly increased order,
in the Experiments Section, we demonstrate that the actual runtime of the proposed method is comparable to  those fastest algorithms.


\section{Regret Analysis}
\label{sec:regret}

We present the high-probability regret upper bound for the proposed {\sf SemiGraphTS} algorithm. A sketch of proof is provided for a key step.  The complete proof can be found in Appendices  \ref{sec:thm42} and \ref{sec:thm41} in the Supplement Material.
%
We assume that the noise term $\eta_{i,j} (t)$ given $\MC{F}_{t-1}$ is $R$-sub-Gaussian, that is, for every $c \in \MB{R}$, 
\BEQ\label{eqn:subGaussian}
\MBE\left[  \exp \{ c \eta_{i,j} (t)  \} \vert \MC{F}_{t-1} \right] \leq \exp (c^2 R^2 / 2), 
\EEQ
for all $i,j,t$, which is a common assumption in the literature for theoretical derivations. 
%
%
The regret bound for {\sf SemiGraphTS}
is described in the following theorem. 

\begin{theorem}\label{thm:regretSemi}
Assume \eqref{eqn:subGaussian} and $\delta \in (0,1)$. Under the semi-parametric linear reward model \eqref{eqn:rewardSemi}, with probability $1 - \delta$, the cumulative regret from {\sf SemiGraphTS} (Algorithm \ref{algo:SemiGraphTS}) achieves
\begin{gather}
R(T) \leq \sum_{j=1}^n 
O \Bigg(\! \Psi_{j,T} \left\{\! \sqrt{d \log (\lvert \MC{T}_{j,T} \rvert)} \!+\! \sqrt{\lambda} \| \Delta_j \| \!\right\}  \times  \label{eqn:thm4.1} \\  
\min \!\left\{\! \sqrt{d \log(dT)}, \! \sqrt{\log (NT)} \!\right\}\!  \sqrt{d \lvert \MC{T}_{j,T} \rvert \log (\lvert \MC{T}_{j,T} \rvert)} \Bigg),  \NN  
\end{gather}
where $\Psi_{j,T} = \sum_{t \in \MC{T}_{j,T}} \| X_t \|_{\Gamma_j(t)^{-1}} / \sum_{t \in \MC{T}_{j,T}} \| X_t \|_{B_j(t)^{-1}}$. 
\end{theorem} 
We note that $\Psi_{j,T} \in (0,1)$ due to $\Gamma_{j_t}(t)^{-1} < B_{j_t}(t)^{-1}$.
A simpler representation of our regret is $\TD{O} (  \max_{j}\!\Psi_{j,T} \cdot d\sqrt{nT} \min\{\sqrt{d}, \sqrt{\log (N)} \} )$, if we assume $\lvert \MC{T}_{j,T} \rvert \approx T/n$ (each $j_t$ is uniformly chosen  at random). 
%
Compared to the regret bound derived in \citet{Yang2020} for the linear graph bandit model, ours has an additional $\min \{ \sqrt{d \log(dT)}, \! \sqrt{\log (NT)} \}$ due to the Thompson sampling; other parts are the same, although our model have additional nonparametric intercept $\nu_j(t)$.
Running  \citet{Kim2019} for each user independently under the same setting  leads to the same form of regret bound with \eqref{eqn:thm4.1}, except for the term $\Psi_{j,T} \{ \sqrt{d \log (\lvert \MC{T}_{j,T} \rvert)} \!+\! \sqrt{\lambda} \| \Delta_j \| \}$ is replaced with  $\sqrt{d \log (\lvert \MC{T}_{j,T} \rvert)} + \sqrt{\lambda}\| \mu_j \|$. 
Since $\Psi_{j,T} \in (0,1)$, the regret bound of the propose algorithm is strictly lower than that from running \citet{Kim2019} independently, provided $\| \Delta_j \| \leq \| \mu_j \|$. 

%

The outline of the proof for Theorem \ref{thm:regretSemi} follows \citet{Agrawal2013} and  \citet{Kim2019}. Major modifications are  made at establishing a high-probability bound for $\WH{\mu}_{j_t}(t) - \mu_{j_t}$, as stated in the theorem below.

\begin{theorem}\label{thm:Emuhat_t_semi}
Assume that the settings for the semi-parametric linear reward model \eqref{eqn:rewardSemi} holds along with \eqref{eqn:subGaussian}.
Let $E^{\WH{\mu}}(t)$ be an event satisfying
\BEQ
E^{\WH{\mu}}(t) = \left\{ \forall i : \lvert b_i^c(t)^T ( \WH{\mu}_{j_t}(t) - \mu_{j_t} ) \rvert \leq s_{i,j_t}^c(t) \alpha(t)  \right\}, \NN
\EEQ
where $b_i^c(t) = b_i(t) - \bar{b}(t)$,  $s_{i,j_t}^c(t) = \| b_i^c(t) \|_{\Gamma_{j_t}(t)^{-1}}$ and
\[
\alpha(t) =
    (4 R + 12) \sqrt{  2 d \log \left\{ \! \frac{24 t^4} { \delta}  \!\left(\!1 + \frac{1}{\lambda} \!\right)\!  \right\} }
    +  \sqrt{2\lambda} (1 + \| \Delta_{j_t} \| ) .
\]
For all $\delta \in (0,1)$ and $t \geq 1$, $\MBP(E^{\WH{\mu}}(t)) \geq 1 - \delta / t^2$.
\end{theorem}

%
The proof for Theorem \ref{thm:Emuhat_t_semi} carefully leverages the structures of $B_j(t)$ and $\Gamma_j(t)$. First, the lemma below enables us to induce $s_{i,j_t}^c(t)$ from $\lvert b_i^c(t)^T ( \WH{\mu}_{j_t}(t) - \mu_{j_t} ) \rvert$ while encapsulating the other terms into quadratic forms associated with $B_{j_t}(t)^{-1}$.
\begin{lemma}\label{lemma:Gamma}
For any $x, y \in \MB{R}^d$ and $j=1,\ldots,n$,
\[
x^T B_j(t)^{-1} y \leq \sqrt{2} \| x \|_{\Gamma_j(t)^{-1}} \| y \|_{B_j(t)^{-1}}.
\]
\end{lemma}
\begin{proof}
For simplicity, let $B_j \!=\! B_j(t)$ and $\Gamma_j \!=\! \Gamma_j(t)$ for all $j$.
By the Cauchy-Schwartz inequality,  $x^T B_j^{-1} y \!=\!  x \Gamma_j^{-\frac{1}{2}} \Gamma_j^{\frac{1}{2}} B_j^{-1} y \!\leq\! \| x \|_{\Gamma_j^{-1}} \sqrt{y^T B_j^{-1} \Gamma_j B_j^{-1} y}$. Note that $B_j^{-1} \!\leq\! (\lambda l_{jj})^{-1} I_d$. Then, by \eqref{eqn:B_jsemi} and \eqref{eqn:Gamma}, $B_j^{-1} \Gamma_j = I_d + \sum_{k \neq j} \lambda^2 l_{jk}^2 B_j^{-1} B_k^{-1} \leq  I_d + \sum_{k \neq j} l_{jk}^2 / (l_{jj} l_{kk}) I_d$. By \eqref{eqn:random_walk_Laplacian}, we have $\sum_{k \neq j}\! l_{jk}^2 / (l_{jj} l_{jk}) \!\leq\! 1$ which yields $B_j^{-1} \Gamma_j \!\leq\! 2 I_d$ and $\sqrt{y^T B_j^{-1} \Gamma_j B_j^{-1} y} \!\leq\! \sqrt{2} \| y\|_{B_j^{-1}}$. This concludes the proof.
\end{proof}
\noindent Then, we utilize  the lemma below to simplify random quadratic forms caused by neighboring users' intermediate estimators $\bar{\mu}_k (t)$ ($k \neq j_t$). 
\begin{lemma}\label{lemma:ineq-Binv}
For any $x \in \MB{R}^d$ and $j,k=1, \ldots,n$,
\[
\| B_k(t)^{-1} x \|_{B_j(t)^{-1}} \leq \| x \|_{B_k(t)^{-1}} / \sqrt{ \lambda^2 l_{jj} l_{kk}}.
\]\end{lemma}
\begin{proof}
By  \eqref{eqn:B_jsemi}, if suffices to show $(uI_d + A) (vI_d + B) (uI_d + A) \geq uv (uI_d + A)$ for any scalars $u,v>0$ and positive semi-definite matrices $A,B$. Observe that
$(uI_d + A) (vI_d + B) (uI_d + A)  = v(uI_d + A)^2 + (uI_d + A)B(uI_d + A)  
\geq v(uI_d + A)^2 
 = u^2 v (I_d + u^{-1}A)^2 
 \geq u^2 v (I_d + u^{-1}A)
 = uv (uI_d + A)$, which completes the proof.
\end{proof}
\noindent Finally, we separately bound each of the simplified terms by employing the technique of \citet{Abbasi-Yadkori2011}. We apply a union bound argument to obtain a uniform bound.  

\begin{proof}[{\bf Sketch of proof for Theorem \ref{thm:Emuhat_t_semi}}] 
 Detailed derivations for key inequalities are provided in Appendix \red{\ref{sec:thm42}} in the Supplementary Material.
Suppose that the semi-parametric reward model \eqref{eqn:rewardSemi} holds.
Fix $t$ and $\delta$. Let $\WH{\mu}_{j_t} (t)$, $B_k(t)$ and $\bar{\mu}_k(t)$ be as in \eqref{eqn:muhat_propsemi}, and \eqref{eqn:mubarsemi}. 
For simplification,
we write as $b_{\tau} = b_{a(\tau)} (\tau)$ and $\eta_{\tau} =\eta_{a(\tau),j_{\tau}} (\tau)$ for $\tau = 1, \ldots, t-1$, and $j=j_t$ with slight abuse of notation.
By algebra and Lemma \ref{lemma:Gamma},
\BEQ\label{eqn:stepAsemi_start}
\lvert b_i^c(t)^T (\WH{\mu}_j (t) - \mu_j) \rvert
\leq \sqrt{2} s_{i,j}^c(t) \sum_{l=1}^6 C_l,
\EEQ
where
{\small
\BEQ
\begin{array}{ll}
C_1 = \| \sum_{k=1}^n \lambda l_{jk} \mu_k \|_{B_j(t)^{-1}}, 
    & C_2 = \| \sum_{k \neq j} \lambda l_{jk} B_k(t)^{-1} \lambda l_{kk}  \mu_k \|_{B_j(t)^{-1}},  \\
C_3 = \| \sum_{\tau \in \MC{T}_{j,t-1}} X_{\tau} \eta_{\tau}  \|_{B_j(t)^{-1}}, 
    & C_4 = \| \sum_{k \neq j} \lambda l_{jk} B_k (t)^{-1} \sum_{\tau \in \MC{T}_{j,t-1}}\!\!\!\! X_{\tau} \eta_{\tau}  \|_{B_j(t)^{-1}},  \\
C_5 = \| A_j (t) \|_{B_j(t)^{-1}}, 
    & C_6 = \| \sum_{k \neq j} \lambda l_{jk} B_k (t)^{-1} A_k (t) \|_{B_j(t)^{-1}}, 
\end{array}
\EEQ
}
%
\noindent with
\[
A_k(t) = \sum_{\tau \in \MC{T}_{k,t\!-\!1}}\!\! D_{\tau} \mu_k + 
\sum_{\tau \in \MC{T}_{k,t\!-\!1}}\!\! 2X_{\tau} \left(\nu_k(\tau)+\bar{b}(\tau)^T\mu_k\right),
\]
$k=1,\ldots,n$,
and
$D_{\tau} = X_{\tau}X_{\tau}^T- \MBE(X_{\tau}X_{\tau}^T \vert \MC{F}_{\tau-1})$. 

For $C_1$, we have $C_1 \leq  \sqrt{\lambda} \| \Delta_{j} \| $  from $B_j(t)^{-1} \leq (\lambda l_{jj})^{-1} I_d$.
For $C_2$,  from Lemma \ref{lemma:ineq-Binv}, we have $C_2 \leq \sqrt{\lambda}   \sum_{k \neq j}   (\lvert  l_{jk} \rvert / \sqrt{l_{jj}})  \| \mu_k \|$ and so $C_2 \leq \sqrt{\lambda}$ by $\| \mu_k \| \leq 1$ and \eqref{eqn:random_walk_Laplacian}.
To bound $C_3$ and $C_4$, we first observe that applying Lemma \ref{lemma:ineq-Binv} to $C_4$ yields
\BEQ\label{eqn:stepA_C3C4_1}
C_3 + C_4 \leq 2\sum_{k =1}^n \frac{\lvert l_{jk} \rvert}{\sqrt{l_{jj} l_{kk}}} \left\| 
  \sum_{\tau \in \MC{T}_{k,t-1}} X_{\tau} \eta_{\tau} \right\|_{B_k(t)^{-1}}.
\EEQ
Next, for each $k$, Lemma A.1 in Appendix \ref{sec:lemma} of the Supplementary Material 
yields the following with probability at least $1 - \delta ( \lvert \MC{T}_{k,t-1} \rvert + 1/n) / 3 t^3$: 
\BEQ\label{eqn:stepA_C3}
\left\| \sum_{\tau \in \MC{T}_{k,t-1}} \!\!\!\! X_{\tau}\eta_{\tau} \right\|_{B_k(t)^{-1}}
  \!\!\!\!\! \leq R \sqrt{ d \log \left\{ \frac{24 t^{4}}{\delta} \left( 1 + \frac{1}{\lambda l_{kk}} \right)   \right\} } . 
\EEQ
\noindent \!\! Since $\sum_{k=1}^n ( \lvert \MC{T}_{k,t-1} \rvert + 1/n ) =  t$, a union bound argument shows that event \eqref{eqn:stepA_C3} holds for all $k=1,\ldots,n$ with probability at least $1 - \delta / t^2$.
Under this event, \eqref{eqn:stepA_C3C4_1} and along with \eqref{eqn:random_walk_Laplacian} yields
\BEQ\label{eqn:stepAsemi_C3C4_2}
C_3 + C_4 \leq   4R \sqrt{ d \log \left\{ \frac{24 t^{4}}{\delta} \left( 1 + \frac{1}{\lambda} \right)   \right\} }. 
\EEQ
Now, for  $C_5$ and $C_6$, applying Lemma \ref{lemma:ineq-Binv} to $C_6$ leads to
\BEQ\label{eqn:stepAsemi_C5C6_1}
C_5 + C_6 
  \leq \sum_{k =1}^n \frac{\lvert l_{jk} \rvert}{\sqrt{l_{jj} l_{kk}}} \left\| A_k (t) \right\|_{B_k(t)^{-1}}.
\EEQ
To bound $\left\| A_k (t) \right\|_{B_k(t)^{-1}}$, we first use the definition for a fixed $k$, 
\BEQ \NN
\left\| A_k (t) \right\|_{B_k(t)^{-1}} \leq
    2 \left\| \sum_{\tau \in \MC{T}_{k,t-1}}\!\!\!\!\!\! X_{\tau} \!\!\left(\nu_k(\tau)+\bar{b}(\tau)^T\mu_k\right) \right\|_{B_k(t)^{-1}}\!\!\!\!  + \left\| \sum_{\tau \in \MC{T}_{k,t-1}} D_{\tau} \mu_k \right\|_{B_k(t)^{-1}}\!\!\!\! .
\EEQ
\noindent \!\!\! Using the fact that $X_{\tau}$ and $D_{\tau}$ are mean-zero random variables given $\MC{F}_{\tau-1}$,
we can follow the techniques in Theorem 4.2 of \citet{Kim2019} to bound each term in the right-hand side of the equation above. Then, by a union bound argument, 
\BEQ\label{eqn:stepAsemi_A_k_2}
\left\| A_k(t) \right\|_{B_k(t)^{-1}}    
  \leq 6 \sqrt{ d \log \left\{ \frac{24 t^4}{\delta}  \left( 1 + \frac{1}{\lambda l_{kk}} \right)   \right\} } 
\EEQ
uniformly for all $k=1,\ldots,n$ with probability at least $1 - 2\delta / (3t^2)$.
Combining \eqref{eqn:stepAsemi_C5C6_1}, \eqref{eqn:stepAsemi_A_k_2} and the definition of random-walk Laplacian \eqref{eqn:random_walk_Laplacian}, we have with probability at least $1 - 2 \delta  / (3t^2)$
\BEQ\label{eqn:stepAsemi_C5C6_2}
C_5 +  C_6 \leq 12 \sqrt{ d \log \left\{ \frac{24 t^{4}}{\delta}   \left( 1 + \frac{1}{\lambda} \right) \right\} }.
\EEQ
Finally, plugging the bounds of $C_1$, $C_2$, \eqref{eqn:stepAsemi_C3C4_2}, and \eqref{eqn:stepAsemi_C5C6_2} into \eqref{eqn:stepAsemi_start} completes the proof.
\end{proof}

\begin{remark}
Our proof used the definition of the random-walk normalized Laplacian to obtain $\sum_{k =1}^n \lvert l_{jk} \rvert /\sqrt{l_{jj} l_{kk}} = 2$. This property does not hold in general in other Laplacian representations; see also \citet{Yang2020} for further discussion.
\end{remark}

\begin{remark}
In deriving the regret bound in Theorem \ref{thm:regretSemi}, we assumed that that $\pi_i(t)$ can be exactly computed, as in \citet{Kim2019}. This assumption appears reasonable since we can choose arbitrary precision to approximate $\pi_i(t)$.
The additional regret caused by the uncertainty of finite Monte Carlo samples 
can be absorbed in the current bound; 
detailed discussion is provided in Appendix \red{\ref{sec:regretMC}} of the Supplementary Material.
\end{remark}

\section{Experiments}
\label{sec:experiments}

We compared the proposed {\sf SemiGraphTS} with algorithms for (i) semi-parametric bandits without exploiting graph, (ii) linear bandits exploiting graph, and (iii) linear bandits without graph. For (i), we included running \citet{Kim2019} independently on $n$ users to fully personalize recommendations (``{\sf SemiTS-Ind}''), running a single instance of \citet{Kim2019} for all users  to synchronize recommendations across users (``{\sf SemiTS-Sin}''). For (ii), we considered 
a Laplacian regularization-based method (\citealt{Yang2020}, namely ``{\sf GraphUCB}'') and clustering-based methods (\citealt{Li2019}, ``{\sf SCLUB}''; \citealt{Li2021}, ``{\sf DyClu}'' ). For (iii), we included ``{\sf LinTS-Ind}'' and ``{\sf LinTS-Sin}'', running \citet{Agrawal2013} in ``independent'' and ``single'' fashions.
 Every bandit algorithm involves a hyperparameter that controls the degree of exploration, either through the variance of $\tilde{\mu}(t)$ in the TS-type algorithms (e.g. $v_j$ in our algorithm) or through the confidence width in the UCB-type algorithms. In graph-based and independent bandit algorithms, we use the same value across users, i.e., $v_j = v$.
Another hyperparameter is $\lambda$, which controls the strength incorporating the graph structure. 
We tuned $(v, \lambda)$ by a grid search for first $t_0$ rounds, with $v \in \{10^{-3}, 10^{-2}, 10^{-1}, 10^{0}, 10^{1}\}$ and $\lambda \in  \{5^{-3}, 5^{-2}, 5^{-1}, 5^{0}, 5^{1}\}$. 
Then, with the best combination of hyperparameters, we assessed each algorithm for over next $T$ rounds.
Other hyperparameters were set as default for each algorithm.
 All computations were conducted in a workstation with AMD Ryzen  3990X CPU and 256GB RAM. All results were generated over five replications. In all Figures, we report the average in solid line and the confidence band (average $\pm \, 1.96 \times \mbox{(standard deviation)}/\sqrt{5}$) in light band.

\paragraph{Synthetic dataset.}
We generated data under \eqref{eqn:rewardSemi}. We considered  $\nu_j(t)$ as $\nu_j(t)=-b_{a*(t)}(t)^T\mu_{j}$ to simulate  a non-stationary scenario and  $\nu_j(t)=0$ for a stationary scenario.  We fixed $n=30, N=10, d=40$. For each time $t$, we chose $j_t$  uniformly  at random.
We constructed  the item features as $b_i(t) = ( I(i\!=\!1) z_{1}(t)^T,  I(i\!=\!2) z_{2}(t)^T, \ldots, I(i\!=\!N) z_{N}(t)^T )^T$, where   $z_i(t)$ follows a uniform distribution on $d'$-dimensional sphere ($d'  = d/N$).  A random error $\eta_{i,j}(t)$ was generated from $\mathcal{N}_d(0,0.1^2)$. 
Next,  the user network $\mathcal{G}$ was generated following the Erd\"os-R\'enyi (ER) model,  in which the edges were generated independently and randomly with probability $p$. We set $p=0.4$. Then we constructed the true user-specific parameters $\mu \in \MB{R}^{nd}$ according to 
$
\mu = \argmin_{ \mu' \in \MB{R}^{nd}} \left[ \left\|\mu' - \mu_0 \right\|^2 + \gamma  \mu'^T ( L \otimes I_d) \mu'  \right],
$
where $\mu_0 \in \MB{R}^{nd}$ is randomly initialized, $L$ is the random-walk graph Laplacian of $\mathcal{G}$, and $\gamma \geq 0$ \citep{Yankelevsky2016}. We put $t_0=5,\!000$ and $T=50,\!000$.
%

\begin{figure}[t]
\centering
    \includegraphics[width=0.98\columnwidth]{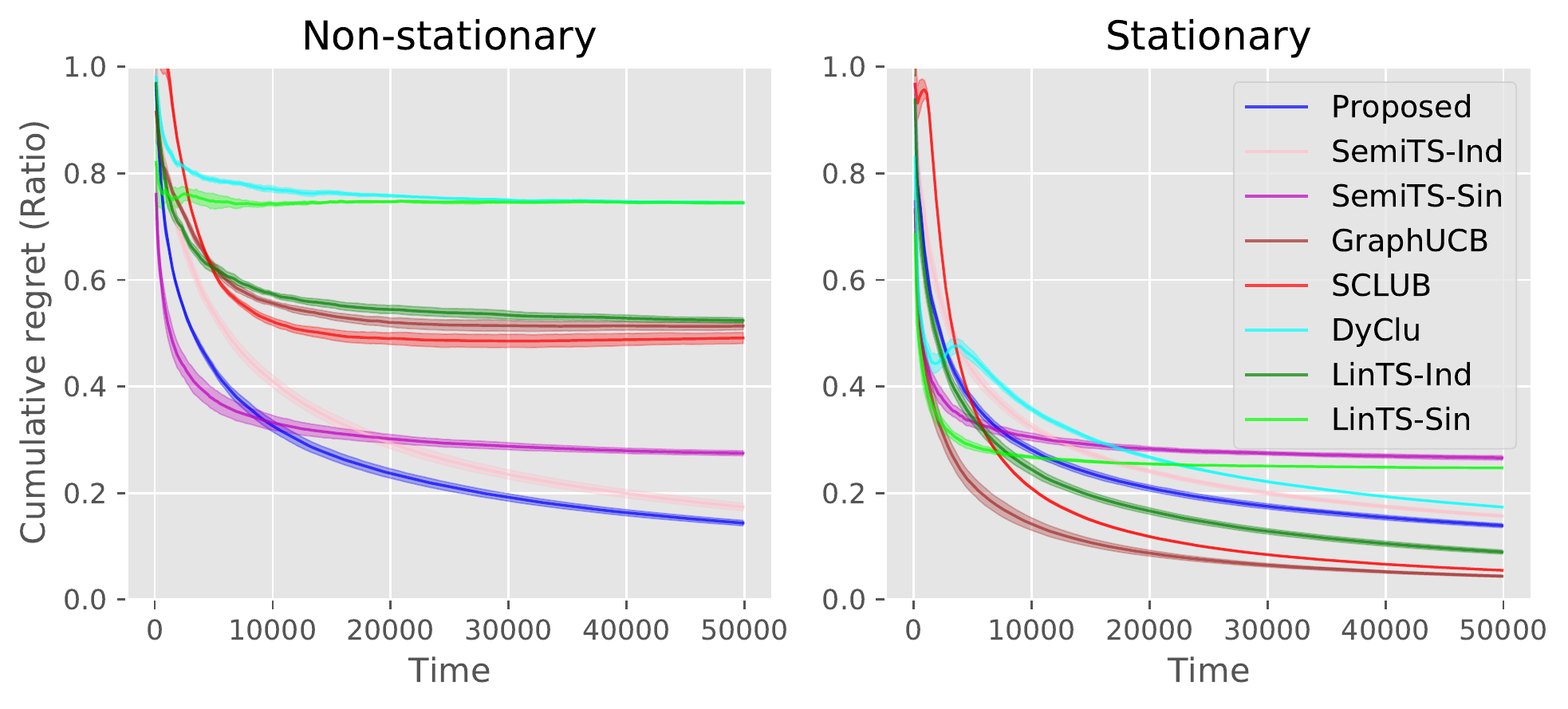}
    \caption{Current cumulative regrets under the non-stationary scenario (left) and the stationary scenario (right). All regrets are relative to that of the random selection.}
\label{fig:1}
\end{figure}

Figure \ref{fig:1} displays the result for the non-stationary scenario with $\gamma=5$. This scenario satisfies all of our assumptions.
As expected, tne proposed {\sf SemiGraphTS} outperformed other algorithms. Compared to {{\sf SemiTS-Ind}} that was the second-best, {\sf SemiGraphTS} additionally exploited the graph structure, which might have led to the final cumulative regret decreased by 11.5 percent. The third best was {{\sf SemiTS-Sin}}, although it performed the best in early rounds. Since {{\sf SemiTS-Sin}} estimates only a small number of parameters, the fitted coefficients may have been converging fast to a biased target.
Another observation is that {{\sf SemiGraphTS}} outperformed the linear graph-based methods. This may suggest that our method could robustly leverage the graph structure when non-stationarity exists. 
As a next experiment, we tested the same setting but under the stationary scenario $\nu_j(t)=0$.
Note that both linear and semi-parametric algorithms have theoretical guarantees for this case. 
The result is reported in  the right panel of Figure \ref{fig:1}. 
We see that the linear graph-based algorithms ({{\sf GraphUCB}} and {\sf SCLUB}) outperformed {{\sf SemiGraphTS}}. Similarly, {{\sf LinTS-Ind}}  outperformed {{\sf SemiTS-Ind}}. 
We hypothesize that accommodating the nuisance terms in semi-parametric algorithms may delay convergence of fitted coefficients, which is a price to pay for robustness.

\begin{figure}[t]
\centering
    \includegraphics[width=0.98\columnwidth]{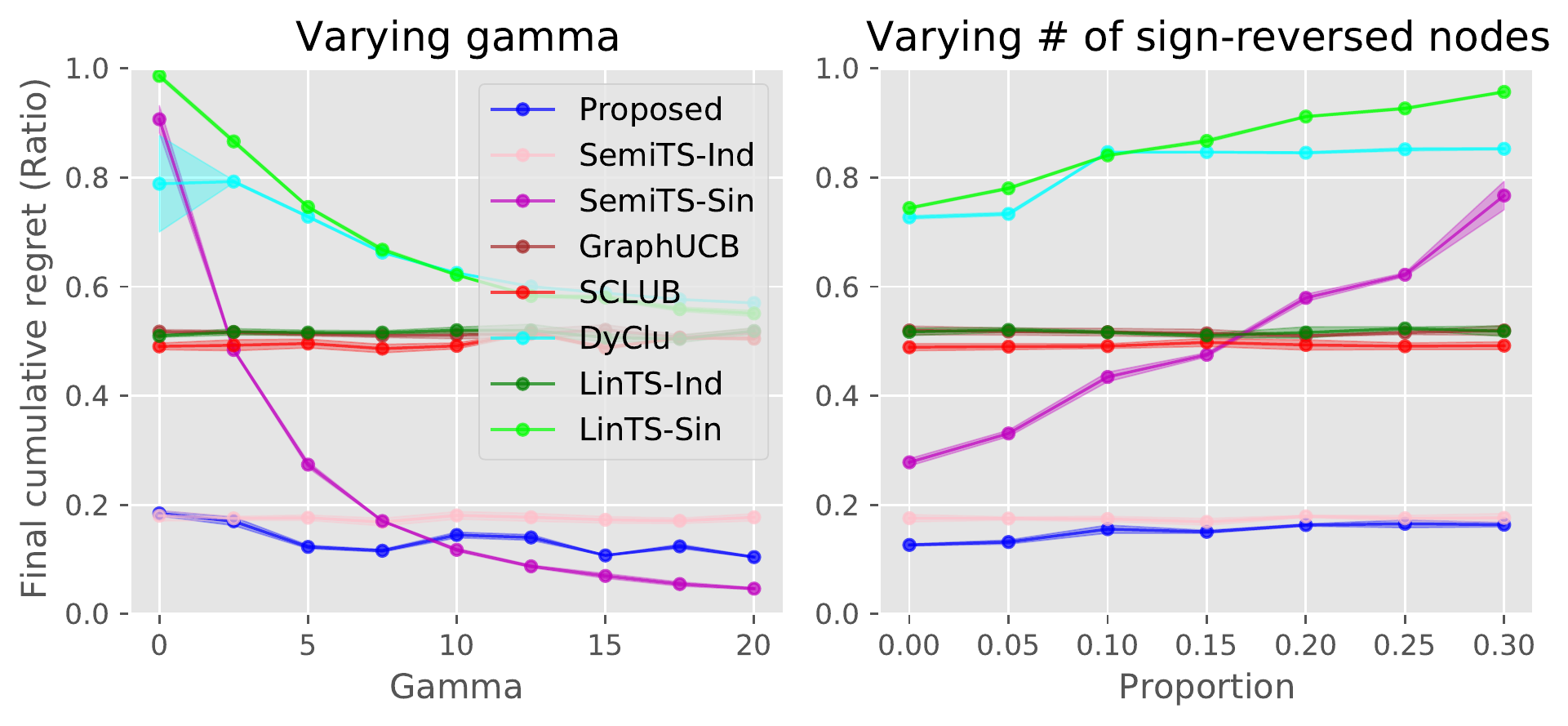}
    \caption{Final cumulative regrets under the non-stationary scenario, while varying $\gamma$ (left) and the proportion of sign-reversed nodes (right). All regrets are relative to that of the random selection.}
    \label{fig:2}
\end{figure}
For sensitivity analysis, we tested the performances of the algorithms against graph strength and graph misspecification. 
In the left panel of Figure \ref{fig:2}, we tracked the final cumulative regrets for varying $\gamma$ from $\gamma = 0$ through $\gamma = 15$, under the non-stationary scenario. A larger $\gamma$ indicates a stronger similarity between $\mu_j$'s. 
For large-$\gamma$ cases, {{\sf SemiGraphTS}} was between those of {{\sf SemiTS-Ind}} and {{\sf SemiTS-Sin}}. For small-$\gamma$ cases, {{\sf SemiGraphTS}} was comparable to {{\sf SemiTS-Ind}} and outperformed {{\sf SemiTS-Sin}} with a large margin. 
The right panel of Figure \ref{fig:2} shows the results for misguided graphs, where we varied the proportion of node $j$s in which the signs of $\mu_j$ were reversed. When the proportion was large, {{\sf SemiGraphTS}} behaved comparably to  {{\sf SemiTS-Ind}}, while {{\sf SemiTS-Sin}} performed poorly.



\begin{figure}[t]
\centering
    \includegraphics[width=0.98\columnwidth]{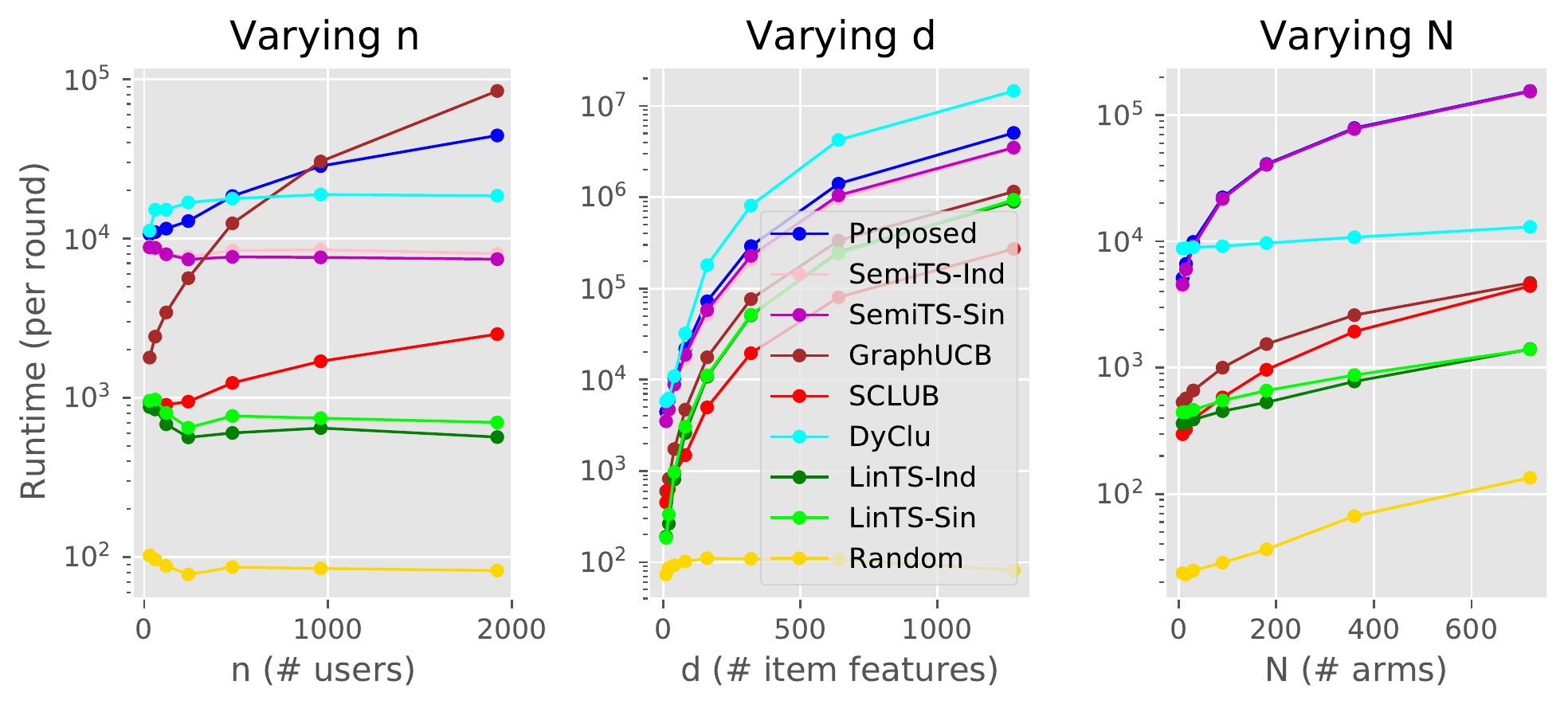}
    \caption{Average runtimes of the algorithms over varying $n$ (left), $d$ (middle), and $N$ (right).}
    \label{fig:3}
\end{figure}

\paragraph{Scalability.}
Figure \ref{fig:3} reports the average runtime per step of each algorithm, varying the number of users $n$ (left panel), the number of features $d$ (middle panel), and the number of arms $N$ (right panel), fixing other settings the same as in the non-stationary synthetic experiment.
{{\sf SemiGraphTS}} was slightly slower than {\sf SemiTS-Ind}. This difference is expected; the construction of $\WH{\mu}_{j_t}(t)$ and $\Gamma_{j_t}(t)$ depends on the degree of the node (user) to serve, which increases linearly with $n$ in the ER graph we tested. 
A comparison of the semi-parametric methods with the linear methods revealed that each of the semi-parametric methods costed more time than its linear counterparts, mainly due to the Monte Carlo approximation of the arm selection probability. One exception was that {\sf SemiGraphTS} was faster than {\sf GraphUCB} as $n$ increases.
Overall, {{\sf SemiGraphTS}} demonstrated comparable efficiency for large graphs when $d$ and $N$ are in a moderate range. 

\paragraph{Real data example.}
The LastFM dataset\footnote{
URLs: \url{https://last.fm/}, \url{http://ir.ii.uam.es/hetrec2011/}} 
is from a music streaming service last.fm, released by \citet{Cantador:RecSys2011}. The dataset consists of $n=1,\!892$ nodes (users) connected by $\lvert E \rvert = 12,\!717$ edges, and $17,\!632$ items (artists) described by $11,\!946$ tags. 
It contains an aggregated table for the frequencies of (user, artist) pairs, representing 
the number of times a user listened to any music of an artist.
We generated an artificial history of  $t_0=5,\!000$ and $T=50,\!000$ rounds following  \citet{Casa-bianchi2013} and \citet{Gentile2014}. 
In short, we randomly sampled one user to serve and $N=25$ artists  for each round.
As item features, we used the first $d=25$ principal component scores resulting from a
term-frequency-inverse-document-frequency (TF-IDF) matrix of artists versus tags, treating artists as ``documents'' and tags as ``words.''
We set the reward to 1 if the selected user ever listened to a selected artist and 0 otherwise.

\begin{figure}[t]
{\centering
    \includegraphics[width=0.75\columnwidth]{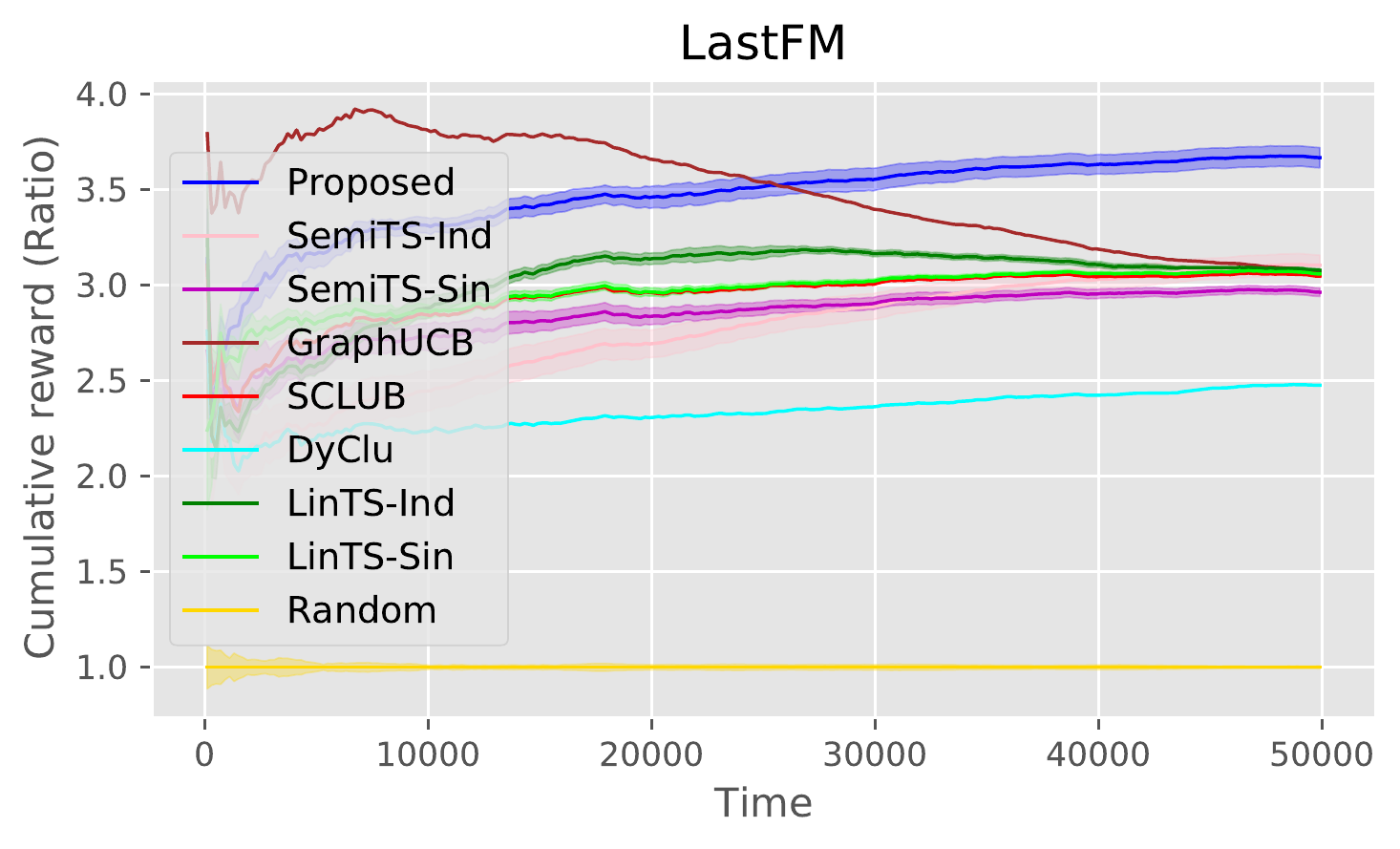}
    \caption{Current cumulative rewards for the LastFM dataset, normalized by the random selection policy.}
    \label{fig:4}
}
\end{figure}

Figure \ref{fig:4} displays the cumulative rewards of the considered algorithms, relative to that of the random selection policy.
{\sf SemiGraphTS} produced the best final cumulative reward, 16.7 percent higher value compared to the second-best algorithms.
In particular, {\sf SemiGraphTS} uniformly outperformed {\sf SemiTS-Ind} and {\sf SemiTS-Sin}, which we believe that the proposed method might have exploited the graph structure successfully. 
Compared to the linear graph-based algorithms, {\sf SemiGraphTS} underperformed {\sf GraphUCB} in early stages but eventually outperformed them.  This result is somewhat anticipated from the synthetic experiment; the presence of nuisance term might have slowed down the learning process of the proposed method but enhanced the robustness of against the change of timely trends.
%
To summary the synthetic and real-data experiments, {Proposed} appears to robustly achieve desirable performances. 


\section{Concluding Remarks}
\label{sec:conclusion}

This study proposes {\sf SemiGraphTS}, the first algorithm for the semi-parametric contextual bandit MAB problem for multiple users equipped with a graph encoding similarity between user preferences. 
{\sf SemiGraphTS} is well suited to more realistic problems in which individual baseline rewards change over time.
Experiments demonstrate the potential advantage of {\sf SemiGraphTS}.

\newpage

\begin{appendices}








\noindent {\Large \bf Supplementary Material}

\bigskip 

In Section \ref{sec:lemma}, we introduce lemmas for theoretical derivation. In Appendix \ref{sec:thm42}, we complete the proof for Theorem \ref{thm:Emuhat_t_semi}. In Appendix \ref{sec:thm41}, we provide the proof for Theorem \ref{thm:regretSemi}. Finally, in Appendix \ref{sec:regretMC}, we discuss the derivation of the regret bound that addresses the approximation to exact $\pi_i(t)$ by Monte Carlo sampling.

\section{Auxiliary Lemmas}\label{sec:lemma}

\begin{lemma}[Simplified version of Corollary 4.3 in \citet{delaPena2004}]\label{lemma:delaPena}
Let $X_{\tau} \in \MB{R}^d$ and $c_{\tau} \in \MB{R}$ be random variables for $\tau = 1, \ldots, t$. Let $A(t) \in \MB{R}^{d \times d}$ be a symmetric and positive semi-definite matrix. Suppose that, for  all $u \in \MB{R}^d$, 
\[
\MBE \left[  \exp \left\{ u^T \sum_{\tau=1}^t X_{\tau} c_{\tau}  -
    \frac{1}{2} u^T A(t) u \right\} \right] \leq 1. 
\]
Then, for any $\delta \in (0,1)$ and any symmetric positive definite matrix $Q \in \MB{R}^{d \times d}$, the following holds with probability at least $1- \delta$:
\[
\left\| \sum_{\tau=1}^t X_{\tau} c_{\tau} \right\|_{(Q + A(t))^{-1}}^2 
    \leq  \log \left\{ \frac{\det(Q + A(t)) / \det(Q)}{\delta^2}  \right\}.
\]
\end{lemma}

The lemma below is Lemma 7 in \citet{delaPena2009}. See also Lemma A.3 of \citet{Kim2019} for proof.

\begin{lemma}\label{lemma:delaPena2}
Let $\{ \MC{F}_{\tau} \}_{\tau=1}^t$ be a filtration. Let $X_{\tau} \in \MB{R}^d$ and $c_{\tau} \in \MB{R}$ be $\MC{F}_{\tau}$-measurable random variables such that $\MBE(X_{\tau} \vert \MC{F}_{\tau-1})=0$, $X_{\tau} \perp c_{\tau} \vert \MC{F}_{\tau-1}$, $\| X_{\tau} \| \leq B$, and $\| c_{\tau} \| \leq 1$ for some constant $B$, $\tau=1, \ldots, t$. Then, for any $u \in \MB{R}^d$,
\[
\MBE \left[  \exp \left\{ u^T \sum_{\tau=1}^t X_{\tau} c_{\tau}  -
    \frac{1}{2} u^T \left( \sum_{\tau=1}^t X_{\tau}X_{\tau}^T + \sum_{\tau=1}^t \MBE (X_{\tau}X_{\tau}^T \vert \MC{F}_{\tau-1}) \right) u \right\} \right] \leq 1. 
\]
\end{lemma}

\begin{lemma}[Azuma-Hoeffding inequality]\label{lemma:AzumaHoeffding}
If $\{ M_t \}_{t=0}^T$ is a supermartingale satisfying $\vert M_t - M_{t-1} \vert \leq c_t$ for all $t$ almost surely, then for any $a > 0$,
\[
\MBP\left( \vert M_T - M_0 \vert \geq a \right) \leq \exp \left( - \frac{a^2}{2 \sum_{t=1}^T c_t^2 } \right).
\]
\end{lemma}

\begin{lemma}[\citealt{Abramowitz1964}]\label{lemma:largedev} If $Z$ is a standard normal random variable, then for any $u \geq 1$,
\[
\frac{1}{2 \sqrt{\pi} u } \exp \left( - \frac{u^2}{2} \right) \leq 
    \MBP \left( \vert Z  \vert > u  \right) \leq 
    \frac{1}{ \sqrt{\pi} u } \exp \left( - \frac{u^2}{2} \right).
\]
\end{lemma}

\section{Proof of Theorem \ref{thm:Emuhat_t_semi}}\label{sec:thm42}

The proof of Theorem \ref{thm:Emuhat_t_semi} follows the sketch in the Regret Analysis Section.

\subsection{Proof of (11)}\label{subsec:thm42_1}

By the semi-parametric reward assumption, for $k=1,\ldots,n$,
\BEA
\bar{\mu}_k(t) &=&  B_k (t)^{-1}  \sum_{\tau \in \MC{T}_{k,t-1}} 
2X_{\tau} \left\{ \nu(\tau)+b_{\tau}^T \mu_k + \eta_{\tau} \right\}  \NN \\ 
&=& 
B_k(t)^{-1}\Bigg\{\sum_{\tau \in \MC{T}_{k,t-1}}2X_{\tau}\nu(\tau)+
\sum_{\tau \in \MC{T}_{k,t-1}} 
 2X_{\tau}  X_{\tau}^T  \mu_k \NN \\
 & &~~~~ +\,\, \sum_{\tau \in \MC{T}_{k,t-1}} 
 2X_{\tau}\bar{b}(\tau)^T\mu_k+
\sum_{\tau \in \MC{T}_{k,t-1}}
2X_{\tau} \eta_{\tau}\Bigg\}  \NN \\
&=& 
B_k (t)^{-1} 
\left\{
 \left( \WH{\Sigma}_{k,t} + \Sigma_{k,t} \right) + 
  \left( \WH{\Sigma}_{k,t} - \Sigma_{k,t} \right)
  + \lambda l_{kk} I_d - \lambda l_{kk} I_d
\right\} \mu_k \NN
 \\
 & &~~~~ +\,\,
B_k (t)^{-1} 
\left\{\sum_{\tau \in \MC{T}_{k,t-1}}
2X_{\tau} \left(\nu(\tau)+\bar{b}(\tau)^T\mu_k\right)+\sum_{\tau \in \MC{T}_{k,t-1}}
2X_{\tau} \eta_{\tau}\right\} \NN \\
&=& 
\mu_k  - \lambda l_{kk} B_k (t)^{-1}  \mu_k  + 
B_k(t)^{-1} A_k(t) +  B_k(t)^{-1}\sum_{\tau \in \MC{T}_{k,t-1}}
2X_{\tau} \eta_{\tau}.  \NN
\EEA

By the relation above, the proposed estimator satisfies
\BEA
&& \WH{\mu}_j (t) - \mu_j \NN \\
&=& - \left[ B_j(t)^{-1} \sum_{k=1}^n \lambda l_{jk} \mu_k \right]
 + \left[ B_j(t)^{-1} \sum_{k \neq j} \lambda^2 l_{jk} l_{kk} B_k(t)^{-1} \mu_k \right]
 \NN \\
& & ~~~+\, 2 \left[ B_j(t)^{-1}\sum_{\tau \in \MC{T}_{k,t-1}}  X_{\tau} \eta_{\tau} \right]
  - 2 \left[ B_j(t)^{-1} \sum_{k \neq j} \lambda l_{jk} 
  B_k(t)^{-1} \sum_{\tau \in \MC{T}_{k,t-1}} X_{\tau} \eta_{\tau}  \right]
 \NN \\
& & ~~~+\,  \left[ B_j(t)^{-1} A_j(t) \right] - \left[ B_j(t)^{-1} \sum_{k \neq j} \lambda l_{jk} B_k(t)^{-1} A_k(t) \right]. \NN
\EEA
Now, left-multiply $b_i^c(t)^T$ on the each side of the equation above and 
applying Lemma \ref{lemma:Gamma} 
on the six terms in the right-hand side yields the desired result.
\qed

\subsection{Proof of \eqref{eqn:stepAsemi_C3C4_2}}\label{subsec:thm42_2}

Fix $k$ ($k=1,\ldots,n$).
Note that when $\MC{F}_{\tau-1}$ and $a(\tau)$ are given, $X_{\tau}$ is fixed and $\eta_{\tau}$ is $R$-sub-Gaussian. Then, from (7), for all $u \in \MB{R}^d$,
\[
\MBE \left[  \exp \left\{ u^T  X_{\tau} \left( \frac{\eta_\tau}{R} \right)  -
    \frac{1}{2} u^T  X_{\tau} X_{\tau}^T u \right\} \Bigg\vert \MC{F}_{\tau-1}, a({\tau}) \right] \leq 1.
\]
This leads to
\[
\MBE \left[  \exp \left\{ u^T \!\!\!\!\sum_{\tau \in \MC{T}_{k,t-1}}\!\!\!\! X_{\tau} c_{\tau}  -
    \frac{1}{2} u^T \WH{\Sigma}_{k,t} u \right\} \right] \leq 1,
\]
which satisfies the assumption of Lemma \ref{lemma:delaPena} with the choice of $X_{\tau} = X_{\tau}$, $c_{\tau} = n_{\tau}/R$,  $Q = \lambda l_{kk} I_d + \Sigma_{k,t}$ and  $A(t) = \WH{\Sigma}_{k,t}$.
Then, for any $0 < \delta < 1$, with probability at least $1 - \delta ( \lvert\MC{T}_{k,t-1}\rvert + 1/n) / (3 t^3)$,
\BEQ\label{eqn:stepA_C3_1}
\left\| \sum_{\tau \in \MC{T}_{k,t-1}} X_{\tau} \eta_{\tau} \right\|_{B_k(t)^{-1}}
  \leq R \sqrt{ \log \left\{ \frac{ \det ( B_k(t) )   / \det (\lambda l_{kk} I_d + \Sigma_{k,t})  }{ (\delta ( \lvert\MC{T}_{k,t-1}\rvert + 1/n) / 3 t^3)^2 } \right\} }. 
\EEQ
We may assume $\lvert\MC{T}_{k,t-1}\rvert \geq 1$, otherwise  the left-hand side of \eqref{eqn:stepA_C3_1} is zero.
The determinant-trace inequality for $\det ( B_k (t) )$ yields
\begin{gather}
\det ( B_k (t) ) \leq \left( \frac{\tr(B_k(t))}{d} \right)^d  \NN \\
 =
 \left( \frac{ \tr( \lambda l_{kk} I_d ) + \sum_{\tau \in \MC{T}_{k,t-1}} \tr( X_{\tau} X_{\tau}^T + \MBE(X_{\tau} X_{\tau}^T \vert \MC{F}_{\tau-1}) ) }{d} \right)^d \NN \\
 \leq
 \left( \lambda l_{kk} + \frac{ 8\lvert\MC{T}_{k,t-1}\rvert }{d} \right)^d, \NN
\end{gather}
where we used $\| X_{\tau} \| \leq 2$.
On the other hand, since  $\Sigma_{k,t}$ is positive semi-definite, we have $ \det (\lambda l_{kk} I_d + \Sigma_{k,t}) \geq  \det (\lambda l_{kk} I_d ) = (\lambda l_{kk})^d$.  Then, for $d \geq 2$ and $t \geq 1$, 
\[
\frac{\det ( B_j (t) )}{ \det (\lambda l_{kk} I_d + \Sigma_{k,t} )  } \leq
    \left( 1 + \frac{ 8\lvert\MC{T}_{k,t-1}\rvert }{d \lambda l_{kk}} \right)^d 
\leq 8^d \lvert\MC{T}_{k,t-1}\rvert^d \left( 1 + \frac{1}{\lambda l_{kk}} \right)^d.
\]
Since $d \geq 2$, $t > 1$, $0 < \delta < 1$ and $1 \leq \lvert\MC{T}_{k,t-1}\rvert \leq t$, 
the right-hand side of  \eqref{eqn:stepA_C3_1} is further simplified by
\begin{align*}
R \sqrt{ \log \left\{ \frac{ \det ( B_k(t) )   / \det (\lambda l_{kk} I_d )  }{ (\delta ( \lvert\MC{T}_{k,t-1}\rvert + 1/n) / 3 t^3)^2 } \right\} }
    & \leq R \sqrt{ \log \left\{ \frac{t^6}{\delta^2} \frac{ 8^d 9 \lvert\MC{T}_{k,t-1}\rvert^d \left( 1 + \frac{1}{\lambda l_{kk}} \right)^d  } { ( \lvert\MC{T}_{k,t-1}\rvert + 1/n )^2 } \right\} } 
    \NN  \\
&\leq R \sqrt{ \log \left\{ \frac{t^{3d}}{\delta^d} \frac{ 8^d 3^d \lvert\MC{T}_{k,t-1}\rvert^d \left( 1 + \frac{1}{\lambda l_{kk}} \right)^d  } { ( \lvert\MC{T}_{k,t-1}\rvert + 1/n )^2 } \right\} } 
    \NN  \\
&\leq R \sqrt{ d \log \left\{ \frac{24 t^{3}}{\delta}  \lvert\MC{T}_{k,t-1}\rvert \left( 1 + \frac{1}{\lambda l_{kk}} \right)   \right\} } 
    \NN \\
&\leq R \sqrt{ d \log \left\{ \frac{24 t^{4}}{\delta} \left( 1 + \frac{1}{\lambda l_{kk}} \right)   \right\} }.
    \NN    
\end{align*}
Combining \eqref{eqn:stepA_C3_1} and the result above, for any $0 < \delta < 1$, we have 
\BEQ\NN
 \left\| \sum_{\tau \in \MC{T}_{k,t-1}} X_{\tau} \eta_{\tau} \right\|_{B_k(t)^{-1}}
  \leq R \sqrt{ d \log \left\{ \frac{24 t^{4}}{\delta} \left( 1 + \frac{1}{\lambda l_{kk}} \right)   \right\} } 
\EEQ
with probability at least $ \geq 1 - \delta ( \lvert\MC{T}_{k,t-1}\rvert + 1/n) / ( 3 t^3)$.
This concludes the derivation.
\qed

\subsection{Proof of \eqref{eqn:stepAsemi_C5C6_1}}\label{subsec:thm42_3}

Fix $k$ ($k=1,\ldots,n$).
Recall the definition of $A_k(t)$,
\BEQ\label{eqn:stepA_C5_1}
\left\| A_k (t) \right\|_{B_k(t)^{-1}} \leq
    2 \left\| \sum_{\tau \in \MC{T}_{k,t-1}} \!\!\!\! X_{\tau} \!\! \left(\nu(\tau)+\bar{b}(\tau)^T\mu_k\right) \right\|_{B_k(t)^{-1}} 
    \!\!\!\!
    +
    \left\| \sum_{\tau \in \MC{T}_{k,t-1}} \!\!\!\! D_{\tau} \mu_k \right\|_{B_k(t)^{-1}} \!\!\!\! .
\EEQ
%
%
For the first term of the right-hand side of \eqref{eqn:stepA_C5_1}, Lemma \ref{lemma:delaPena2} yields
\BEQ\NN
\MBE \left[  \exp \left\{ u^T \sum_{\tau \in \MC{T}_{k, t-1}} X_{\tau} c_{\tau}  -
    \frac{1}{2} u^T \left( \WH{\Sigma}_{k,t} + \Sigma_{k,t} \right) u \right\} \right] \leq 1
\EEQ
for any $u \in \MB{R}^d$,
where $c_{\tau} =  (\nu(\tau)+\bar{b}(\tau)^T\mu_k) / 2$. Then, we can apply Lemma \ref{lemma:delaPena} with $A(t) = \WH{\Sigma}_{k,t} + \Sigma_{k,t}$ and $Q = \lambda l_{kk} I_d$ to obtain the following inequality  with probability at least $1 - \delta ( \lvert\MC{T}_{k,t-1}\rvert + 1/n) / 3t^3$:
\BEQ \label{eqn:stepA_C5_2}
\left\| \sum_{\tau \in \MC{T}_{k,t-1}} X_{\tau} \left(\nu(\tau)+\bar{b}(\tau)^T\mu_k\right) \right\|_{B_k(t)^{-1}}
  \leq 2 \sqrt{ \log \left\{ \frac{ \det ( B_k(t) )   / \det (\lambda l_{kk} I_d)  }{ (\delta ( \lvert\MC{T}_{k,t-1}\rvert + 1/n) / 3t^3)^2 } \right\} }. 
\EEQ
We can bound \eqref{eqn:stepA_C5_2} similarly as in bounding the right-hand side of \eqref{eqn:stepA_C3_1}. Therefore, with probability at least $1 - \delta ( \lvert\MC{T}_{k,t-1}\rvert + 1/n) / 3t^3$,
\BEQ\label{eqn:stepA_C5_2_1}
\left\| \sum_{\tau \in \MC{T}_{k,t-1}} X_{\tau} \left(\nu(\tau)+\bar{b}(\tau)^T\mu_k\right) \right\|_{B_k(t)^{-1}}
  \leq  2 \sqrt{ d \log \left\{ \frac{24 t^{4}}{\delta}   \left( 1 + \frac{1}{\lambda l_{kk}} \right) \right\} }. 
\EEQ
%
%
For the second term of the right-hand side of \eqref{eqn:stepA_C5_1}, we let $Y_{k,\tau} = D_{\tau} \mu_k$ and observe $Y_{k,\tau} \in \MB{R}^d$,  $\MBE(Y_{k,\tau} \vert \MC{F}_{\tau-1}) = 0$. It is straightforward from Lemma 4.4 and its proof in \citet{Kim2019} to derive
\[
\MBE \left[  \exp \left\{ u^T \sum_{\tau \in \MC{T}_{k, t-1}} \frac{1}{\sqrt{2}} Y_{k,\tau} -
    \frac{1}{2} u^T \left( \WH{\Sigma}_{k,t} + \Sigma_{k,t} \right) u \right\} \right] \leq 1
\]
for any $u \in \MB{R}^d$,
which again satisfies the assumption of Lemma \ref{lemma:delaPena} with the choice of $X_\tau = Y_{k,\tau}$, $c_\tau = 1/\sqrt{2}$ and $A(t) = \WH{\Sigma}_{k,t} + \Sigma_{k,t}$. Then, putting $Q = \lambda l_{kk} I_d$, we have with probability at least $1 - \delta ( \lvert\MC{T}_{k,t-1}\rvert + 1/n) / 3t^3$,
\BEA
\left\| \sum_{\tau \in \MC{T}_{k,t-1}} D_{\tau} \mu_k \right\|_{B_k(t)^{-1}}
  &\leq& \sqrt{2} \sqrt{ \log \left\{ \frac{ \det ( B_k(t) )   / \det (\lambda l_{kk} I_d)  }{ (\delta ( \lvert\MC{T}_{k,t-1}\rvert + 1/n) / 3t^3)^2 } \right\} } \NN \\
&\leq& \sqrt{2} \sqrt{ d \log \left\{ \frac{24 t^{4}}{\delta}   \left( 1 + \frac{1}{\lambda l_{kk}} \right) \right\} }. \label{eqn:stepA_C5_3}
\EEA
Plugging \eqref{eqn:stepA_C5_2_1} and \eqref{eqn:stepA_C5_3} into \eqref{eqn:stepA_C5_1} yields bounds for each user:
\[
\MBP \left[ \left\| A_k (t) \right\|_{B_k(t)^{-1}} \leq 6 \sqrt{ d \log \left\{ \frac{24 t^{4}}{\delta}   \left( 1 + \frac{1}{\lambda l_{kk}} \right) \right\} }
\right] \geq  1 - \frac{2 \delta ( \lvert\MC{T}_{k,t-1}\rvert + 1/n)}{ 3t^3 }.
\]
Finally, applying the union bound argument yields
\BEQ\label{eqn:stepA_C5_4}
\MBP \left[ \forall k =1,\ldots,n \,:\, \left\| A_k(t) \right\|_{B_k(t)^{-1}}    
  \leq 6 \sqrt{ d \log \left\{ \frac{24t^{4}}{\delta} \left( 1 + \frac{1}{\lambda l_{kk}} \right)   \right\} } 
  \right] \geq 1 - \frac{2 \delta} { 3t^2 },
\EEQ
which completes the proof.
\qed

\section{Proof of Theorem \ref{thm:regretSemi}}\label{sec:thm41}


The proof incorporates the lines of \citet{Agrawal2013} and \citet{Kim2019} with the proposed estimation and Thompson sampling steps.
Throughout the Section, we write as $j=j_t$, $b_{\tau} = b_{a(\tau)} (\tau)$ and $\eta_{\tau} =\eta_{a(\tau),j_{\tau}} (\tau)$ for brevity. We reserve $k$ ($k=1,\ldots,k$) to denote user index.
The proof has six steps:
\BIT
\item[(a)] (Theorem \ref{thm:Emuhat_t_semi}) To establish a high-probability upper bound of $\lvert b_i^c(t) (\WH{\mu}_j (t) - \mu_j) \rvert$.
\item[(b)] (Lemma \ref{lemma:stepB_uncentered}) To establish a high-probability upper bound of $\lvert b_i^c(t) (\TD{\mu}_j (t) - \WH{\mu}_j(t)) \rvert$ given $\MC{F}_{t-1}$.
\item[(c)] (Definition \ref{def:C_t_uncentered}) To divide arms at each time $t$ into saturated arms and unsaturated arms.
\item[(d)] (Lemma \ref{lemma:stepD_uncentered}) To bound the probability of playing saturated arms by a function of playing unsaturated arms.
\item[(e)] (Lemma \ref{lemma:stepE_uncentered}) To bound $regret(t)$ given $\MC{F}_{t-1}$ for each $t$.
\item[(f)] To bound $R(T)$ and complete the proof.
\EIT

We begin with step (b).

\begin{lemma}\label{lemma:stepB_uncentered}
Let $E^{\TD{\mu}}(t)$ be an event defined by
\BEQ\NN
E^{\TD{\mu}}(t) = \left\{ \forall i : \lvert b_i^c(t)^T ( \TD{\mu}_j(t) - \WH{\mu}_j(t) ) \rvert \leq 
    v_j s^c_{i,j}(t)  \min \{ \sqrt{4 d \log (2dT)}, \sqrt{4 \log (2NT)} \}    \right\}.
\EEQ
for all  $t \geq 1$, $\MBP(E^{\TD{\mu}}(t) \vert \MC{F}_{t-1}) \geq 1 - 1 / T^2$.
\end{lemma}
\begin{proof}
We first show  $\lvert b_i^c(t)^T ( \TD{\mu}_j(t) - \WH{\mu}_j(t) ) \rvert \leq v_j  s^c_{i,j}(t) \sqrt{4  d \log (2dT)}$.
Given $\MC{F}_{t-1}$, the values of $b_i^c(t)$, $\Gamma_j(t)$, and $\WH{\mu}_j(t)$ are fixed.
Then, for $i=1, \ldots, N$, we  have
\begin{align}
\left\vert b_i^c(t)^T ( \TD{\mu}_j(t) - \WH{\mu}_j(t) ) \right\vert 
&= \left\vert v_j b_i^c(t)^T \Gamma_j(t)^{-\frac{1}{2}} \cdot \frac{1}{v_j} \Gamma_j(t)^{\frac{1}{2}} ( \TD{\mu}_j(t) - \WH{\mu}_j(t) ) \right\vert    \NN \\
&\leq v_j s^c_{i,j}(t) \left\| \frac{1}{v_j} \Gamma_j(t)^{\frac{1}{2}} ( \TD{\mu}_j(t) - \WH{\mu}_j(t) ) \right\|_2 
\NN \\
&= v_j s^c_{i,j}(t) \sqrt{\sum_{l=1}^d Z_l(t)^2 }, 
\label{eqn:stepB_uncentered_1}
\end{align}
where $Z_l(t) \vert \MC{F}_{t-1}$ $(l = 1, \ldots, d)$ identically and independently follow the standard normal distribution. We apply Lemma \ref{lemma:largedev} with the choice of $u = \sqrt{2 \log (2dT^2)}$. Noting $\sqrt{2 \log (2 dT^2)} \leq \sqrt{2 \log (2^2d^2T^2)} = \sqrt{4 \log (2dT)}$,
\begin{gather}
\MBP \left( \lvert Z_l(t) \rvert > \sqrt{4 \log (2dT)} \big\vert \MC{F}_{t-1} \right)
 \leq \MBP \left( \lvert Z_l(t) \rvert > \sqrt{2 \log (2dT^2)} \big\vert \MC{F}_{t-1} \right) \NN \\
 \leq \frac{1}{ \sqrt{ 2 \pi \log (2dT^2)}} \cdot \frac{1}{2dT^2}  
 \leq \frac{1}{2dT^2}, \NN
\end{gather}
for each $l=1,\ldots,d$.
Then, by a union bound argument,
\BEQ \label{eqn:stepB_uncentered_2}
\MBP \left( \forall l=1,\ldots,d \,:\, \lvert Z_k(t) \rvert > \sqrt{4 \log (2dT)} \big\vert \MC{F}_{t-1} \right)
 \leq \frac{1}{2T^2}.
\EEQ
Therefore, combining \eqref{eqn:stepB_uncentered_1} and \eqref{eqn:stepB_uncentered_2} yields
\[
\MBP \left( \forall i :
\left\vert b_i^c(t)^T ( \TD{\mu}_j(t) - \WH{\mu}_j(t) ) \right\vert 
\leq v_j  s^c_{i,j}(t) \sqrt{4  d \log (2dT)}
\right) \geq 1 - \frac{1}{2 T^2}.
\]
On the other hand, by the observation that $b_i^c(t)^T ( \TD{\mu}_j(t) - \WH{\mu}_j(t) )  \vert \MC{F}_{t-1}$ $(i = 1, \ldots, N)$ identically and independently follow the standard normal distribution,
one can apply a similar technique to derive $\lvert b_i^c(t)^T ( \TD{\mu}_j(t) - \WH{\mu}_j(t) ) \rvert \leq v_j  s^c_{i,j}(t) \sqrt{4  \log (2NT)}$ with probability at least $1 - 1/(2T^2)$ given $\MC{F}_{t-1}$. Combining the two bounds, we obtain the desired result.
\end{proof}

In step (c), we divide arms at each time $t$ into saturated arms and unsaturated arms. Note that $C(t)$ implitly depends on $j_t$. 
\begin{definition}\label{def:C_t_uncentered}
Define  $C(t)$, the set of saturated arms, by
\[
C (t) = \{ i : b_i^c(t)^T \mu_j + g_j(T) s^c_{i,j}(t) < b_{a^* (t)}(t)^T \mu_j \},
\]
where $g_k(T) = \alpha_k(T) + v_k \min \{ \sqrt{4d \log(2dT)}, \sqrt{4 \log(2NT)} \}$ and $\alpha_k(T) = (4 R + 12) \cdot$ $\sqrt{  d \log \left\{ (24 T^4  / \delta ) (1 + \lambda^{-1} )  \right\} }   +  \sqrt{\lambda} (1 + \| \Delta_{k} \| )$, $k=1,\ldots,n$.
\end{definition}

In step (d), we establish that the probability of playing saturated arms is bounded by the probability of playing unsaturated arms up to constant multiplication and addition.
\begin{lemma}\label{lemma:stepD_uncentered}
Given $\MC{F}_{t-1}$ such that $E^{\WH{\mu}}(t)$ is true, 
\[
\MBP\left( a(t) \in C(t) \vert \MC{F}_{t-1} \right) \leq
\frac{1}{p} \MBP\left( a(t) \notin C(t) \vert \MC{F}_{t-1} \right)
+ \frac{1}{pT^2},
\]
where $p = 1/(4e \sqrt{\pi})$.
\end{lemma}
\begin{proof}
Since $a(t) = \argmax_{1 \leq i \leq N} \{ b_i^c(t)^T \TD{\mu}_j(t) \}$ by definition, if $b_{a^*(t)}(t)^T \TD{\mu}_j(t) > b_i^c(t)^T \TD{\mu}_j(t)$ for every $i \in C(t)$, then $a(t) \notin C(t)$. This implies
\BEQ\label{eqn:stepD_uncentered_1}
\MBP\left( a(t) \notin C(t) \vert \MC{F}_{t-1} \right) \geq
\MBP\left( \forall i \in C(t) : b_{a^*(t)}(t)^T \TD{\mu}_j(t) > b_i^c(t)^T \TD{\mu}_j(t)  \vert \MC{F}_{t-1} \right).
\EEQ
On the other hand, when $E^{\TD{\mu}}(t)$ is additionally true, 
\begin{align*}
b_i^c(t)^T \TD{\mu}_j(t) 
&\leq b_i^c(t)^T \mu_j + g_j(T) s^c_{i,j}(t) &\mbox{(Def. of $E^{\WH{\mu}}(t)$ \& $E^{\TD{\mu}}(t)$)} \NN \\
&\leq b_{a^*(t)}(t)^T \mu_j, &\mbox{(Def. of $C(t)$)}, \NN
\end{align*}
which implies that
\begin{gather}
\MBP\left( b_{a^*(t)}(t)^T \mu_j < b_{a^*(t)}(t)^T \TD{\mu}_j(t)  \vert \MC{F}_{t-1} \right) \NN \\
\leq 
\MBP\left(  \forall i \in C(t) :  b_i^c(t)^T \TD{\mu}_j(t) < b_{a^*(t)}(t)^T \TD{\mu}_j(t) \vert \MC{F}_{t-1} \right) + \left( 1 - \MBP\left( E^{\TD{\mu}}(t) \vert \MC{F}_{t-1} \right) \right)
\label{eqn:stepD_uncentered_2}
\end{gather}
The left-hand side of \eqref{eqn:stepD_uncentered_2} can be lower-bounded, because the normality of $\TD{\mu}_j(t)$ and Lemma \ref{lemma:largedev} yields
\begin{align*}
& \MBP\left( b_{a^*(t)}(t)^T \TD{\mu}_j(t) > b_{a^*(t)}(t)^T \mu_j \Big\vert \MC{F}_{t-1} \right)  \NN \\
&=
\MBP\left( \frac{b_{a^*(t)}(t)^T ( \TD{\mu}_j(t) - \WH{\mu}_j(t) ) }{v_j s^c_{a^*(t), j}(t)} > \frac{b_{a^*(t)}(t)^T ( \mu_j - \WH{\mu}_j(t) ) }{v_j s^c_{a^*(t), j}(t)} \Big\vert \MC{F}_{t-1} \right) \NN \\
&\geq \MBP \left( Z(t) > \frac{\alpha_j(T)}{v_j} \big\vert \MC{F}_{t-1} \right) \NN \\
&\geq \frac{1}{4 \sqrt{\pi} u} \exp\left( - \frac{u^2}{2} \right), \NN
\end{align*}
where $u = \alpha_j(T)/v_j$ and $Z(t)\vert\MC{F}_{t-1}$ is a standard normal random variable. 
Note that $u \leq 1$ by the construction. Therefore,

\BEQ\label{eqn:stepD_uncentered_3}
\MBP\left( b_{a^*(t)}(t)^T \TD{\mu}_j(t) > b_{a^*(t)}(t)^T \mu_j \vert \MC{F}_{t-1} \right) \geq \frac{1}{4 e \sqrt{\pi}} = p.
\EEQ
Combining \eqref{eqn:stepD_uncentered_1}, \eqref{eqn:stepD_uncentered_2}, \eqref{eqn:stepD_uncentered_3} and Lemma \ref{lemma:stepB_uncentered}, we have
\[
\MBP\left( a(t) \notin C(t) \vert \MC{F}_{t-1} \right) + \frac{1}{T^2} \geq p,
\]
which implies
\[
\frac{\MBP\left( a(t) \in C(t) \vert \MC{F}_{t-1} \right) }{\MBP\left( a(t) \notin C(t) \vert \MC{F}_{t-1} \right) + \frac{1}{T^2}} \leq \frac{1}{p}.
\]
This completes the proof.
\end{proof}

Before proceeding to bound the cumulative regret, we bound each $regret(t)$ given $\MC{F}_{t-1}$ in step (e).

\begin{lemma}\label{lemma:stepE_uncentered}
Given $\MC{F}_{t-1}$ such that $E^{\WH{\mu}}(t)$ is true, 
\[
\MBE\left( regret(t) \vert \MC{F}_{t-1} \right)
\leq \frac{5 g_j(T)}{p} \MBE\left( s^c_{a(t), j}(t) \vert \MC{F}_{t-1} \right)
+ \frac{4 g_j(T)}{pT^2}.
\]
\end{lemma}
\begin{proof}
Let $\bar{a}(t) = \argmin_{i \notin C(t)} s^c_{i,j}(t)$. If $\MC{F}_{t-1}$ is given, then $\bar{a}(t)$ is deterministic.  This value is also well-defined due to $a^*(t) \notin C(t)$. Under $\MC{F}_{t-1}$ such that both $E^{\WH{\mu}}(t)$ and $E^{\TD{\mu}}(t)$ holds,
\begin{align*}
& b_{a^*(t)}(t)^T \mu_j \NN \\
&= b_{a^*(t)}(t)^T \mu_j - b_{\bar{a}(t)}(t)^T \mu_j + b_{\bar{a}(t)}(t)^T \mu_j \NN \\
&\leq g_j(T) s^c_{\bar{a}(t), j}(t) + b_{\bar{a}(t)}(t)^T \mu_j
    &\mbox{($\bar{a}(t) \notin C(t)$ \& def. of $C(t)$)} \NN \\
&\leq g_j(T) s^c_{\bar{a}(t), j}(t) + b_{\bar{a}(t)}(t)^T \TD{\mu}_j(t)
    + g_j(T) s^c_{\bar{a}(t), j}(t)
    &\mbox{(Def. of $E^{\WH{\mu}}(t)$ and $E^{\TD{\mu}}(t)$)} \NN \\
&\leq 2g_j(T) s^c_{\bar{a}(t), j}(t) +  b_{a(t)}(t)^T \TD{\mu}_j(t)
    &\mbox{(Def. of $a(t)$)} \NN \\
&\leq 2g_j(T) s^c_{\bar{a}(t), j}(t) + b_{a(t)}(t)^T \mu_j
    + g_j(T) s^c_{a(t), j}(t) 
    &\mbox{(Def. of $E^{\WH{\mu}}(t)$ and $E^{\TD{\mu}}(t)$)}, \NN
\end{align*}
which yields
\BEQ\NN\label{eqn:stepE_uncentered_1}
regret(t) \leq 2g_j(T) s^c_{\bar{a}(t), j}(t) + g_j(T) s^c_{a(t), j}(t).
\EEQ
Then, under $\MC{F}_{t-1}$ such that $E^{\WH{\mu}}(t)$ holds, the following holds from inequality above, Lemma \ref{lemma:stepB_uncentered} and $\lvert regret(t) \rvert \leq 2$:
\BEA
&& \MBE\left( regret(t) \vert \MC{F}_{t-1} \right)  \NN \\
    &=& \MBE\left( regret(t) I(E^{\TD{\mu}}(t)) \vert \MC{F}_{t-1} \right)
        + \MBE\left( regret(t) \{1 - I(E^{\TD{\mu}}(t)) \} \vert \MC{F}_{t-1} \right) \NN \\
    &\leq& 2g_j(T) s^c_{\bar{a}(t), j}(t) + 
    g_j(T) \MBE\left( s^c_{a(t), j}(t) \vert \MC{F}_{t-1} \right)
    + 2 \left( 1 - \MBP\left( E^{\TD{\mu}}(t) \vert \MC{F}_{t-1} \right) \right) \NN \\
&\leq& 2g_j(T) s^c_{\bar{a}(t), j}(t) + 
    g_j(T) \MBE\left( s^c_{a(t), j}(t) \vert \MC{F}_{t-1} \right)
    + \frac{2}{T^2}.   \label{eqn:stepE_uncentered_2}
\EEA

We now further bound $s^c_{\bar{a}(t), j}(t)$. Observe that
\begin{align*}
& s^c_{\bar{a}(t), j}(t) \\
&= s^c_{\bar{a}(t), j}(t) \left\{ \MBP( a(t) \in C(t) \vert \MC{F}_{t-1} ) + \MBP( a(t) \notin C(t) \vert \MC{F}_{t-1} ) \right\} \\
&= s^c_{\bar{a}(t), j}(t) \left\{ \frac{2}{p} \MBP( a(t) \notin C(t) \vert \MC{F}_{t-1} ) + \frac{1}{pT^2} \right\} & \mbox{(Lemma \ref{lemma:stepD_uncentered})} \\
&=  \frac{2}{p} \MBE\left( s^c_{\bar{a}(t), j}(t) I( a(t) \notin C(t)) \vert \MC{F}_{t-1} \right)
    + \frac{s^c_{\bar{a}(t), j}(t)}{pT^2} \\
&\leq \frac{2}{p} \MBE\left( s^c_{a(t), j}(t) I( a(t) \notin C(t)) \vert \MC{F}_{t-1} \right)
    + \frac{s^c_{\bar{a}(t), j}(t)}{pT^2} & \mbox{(Def. of $\bar{a}(t)$)} \\
&\leq  \frac{2}{p} \MBE\left( s^c_{a(t), j}(t) \vert \MC{F}_{t-1} \right)
    + \frac{1}{pT^2}. & \mbox{($s^c_{i,k}(t) \leq 1$ for any $i,k,t$)}
\end{align*}
Combining the inequality above and \eqref{eqn:stepE_uncentered_2} conclude the proof.
\end{proof}

In step (f), we complete the proof.
\begin{proof}[{\bf Proof for Theorem \ref{thm:regretSemi}}] 
Let 
\[
M_t := regret(t) I( E^{\WH{\mu}}(t)) - \frac{5g_{j_t}(T)}{p} s_{a(t), j_t}(t) - \frac{4g_{j_t}(T)}{pT^2},  ~ t=1, \ldots, T,
\]
with $M_0 = 0$. 

We apply martingale arguments for each user $k=1,\ldots,n$, and aggregate them by union bound. Fix $k$ and let $T_k =\lvert \MC{T}_{k,T} \rvert$.
Due to Lemma \ref{lemma:stepE_uncentered} and $s^c_{i,k}(t) \leq 1$, 
$\{ M_t \}_{t \in \{0\}\cup\MC{T}_{k,T}}$ is a supermartingale process satisfying  $\lvert M_t \rvert \leq 10 g_k(T) /p$. We apply Lemma \ref{lemma:AzumaHoeffding} with the choice of $c_t = 10 g_k(T) /p$ and $a = (10g_k(T)/p) \sqrt{ 2 T_k \log(2T/(\delta T_k))}$ that satisfies $\exp ( - a^2 / (2 \sum_t c_t^2) ) = \delta T_k / (2T)$.
This yields
\begin{gather}
\sum_{t \in \MC{T}_{k,T}} regret(t) I( E^{\WH{\mu}}(t))  \leq
\NN \\
 \frac{5g_k(T)}{p} \sum_{t \in \MC{T}_{k,T}} s^c_{a(t), k}(t) + \frac{4g_k(T)}{pT}
 + \frac{10g_k(T)}{p} \sqrt{ 2T_k \log(2 T / (\delta T_k) )}
\end{gather}
with probability at least $1 - \delta T_k / (2T)$. Since $T_1 + \ldots + T_n = T$, a union bound argument over $k=1,\ldots,n$ leads to
\begin{gather}
\sum_{t=1}^T regret(t) I( E^{\WH{\mu}}(t)) \leq \NN \\
 \sum_{k=1}^n \left[ \frac{5g_k(T)}{p} \sum_{t \in \MC{T}_{k,T}} s_{a(t), k}(t) + \frac{4g_k(T)}{pT}
 + \frac{10g_k(T)}{p} \sqrt{ 2T_k \log(2 T / (\delta T_k) )} \right]    
\end{gather}
with probability at least $1 - \delta / 2$.

On the other hand,  we apply a union bound argument to Theorem \ref{thm:Emuhat_t_semi} over $t=1,\ldots,T$ and replace  $\delta$ with $ 3\delta/\pi^2$, which yields $\MBP ( E^{\WH{\mu}}(t)  \mbox{ for all $t=1, \ldots, T$}) \geq 1 - \delta / 2$.
Then, $regret(t) I( E^{\WH{\mu}}(t)) = regret(t)$ for every $t$ with probability at least  $1 - \delta / 2$. 

Therefore, with probability at least $1 - \delta$,
{\small
\[
\sum_{t=1}^T regret(t)  \leq
 \sum_{k=1}^n \left[ \frac{5g_k(T)}{p} \!\! \sum_{t \in \MC{T}_{k,T}} \!\! s^c_{a(t), k}(t) + \frac{4g_k(T)}{pT}
 + \frac{10g_k(T)}{p} \sqrt{ 2T_k \log(2 T / (\delta T_k) )} \right].
\]
}
Now, by Lemma \ref{lemma:covsum} below and the definitions of $g_k(T)$ and $p$, 
\begin{gather}
R(T) \leq \sum_{k=1}^n 
O \Bigg(\! \Psi_{k,T}  \left\{\! \sqrt{d \log (\lvert \MC{T}_{k,T} \rvert)} \!+\! \sqrt{\lambda} \| \Delta_k \| \!\right\}\! \times     \NN \\
\min \!\left\{\! \sqrt{d \log(d\lvert \MC{T}_{k,T} \rvert)}, \! \sqrt{\log (N\lvert \MC{T}_{k,T} \rvert)} \!\right\}\! 
\sqrt{d\lvert \MC{T}_{k,T} \rvert \log (\lvert \MC{T}_{k,T} \rvert)} \Bigg)    \NN
\end{gather}
with probability at least $1 - \delta$, which completes the proof.
\end{proof}

\begin{lemma}\label{lemma:covsum}
\[
\sum_{t \in \MC{T}_{k,T}} s^c_{a(t), k} (t) =
 O \left( \Psi_{k,T} \sqrt{d\lvert \MC{T}_{k,T} \rvert \log(\lvert \MC{T}_{k,T} \rvert)} \right).
\]
\end{lemma}
\begin{proof}
We recall that $\Psi_{k,T} = \sum_{t \in \MC{T}_{k,T}} \| X_t \|_{\Gamma_k(t)^{-1}} / \sum_{t \in \MC{T}_{k,T}} \| X_t \|_{B_k(t)^{-1}}$ and that $\Psi_{k,T}  \in (0,1)$ due to $\Gamma_k(t)^{-1} < B_k(t)^{-1}$ for all $j$, $t$.
Since $\sum_{t \in \MC{T}_{k,T}} s^c_{a(t), k} (t) =  \sum_{t \in \MC{T}_{k,T}} \| X_t \|_{\Gamma_k(t)^{-1}}$ by the definitions of $s^c_{i, k}(t)$ and $X_t$, we have
\[
\sum_{t \in \MC{T}_{k,T}} s^c_{a(t), k} (t) =
  \Psi_{k,T} \sum_{t \in \MC{T}_{k,T}} \| X_{\tau} \|_{B_k(t)^{-1}}.
\]
We now claim $\sum_{t \in \MC{T}_{k,T}} \| X_{\tau} \|_{B_k(t)^{-1}} = O \left(\sqrt{d\lvert \MC{T}_{k,T} \rvert \log(\lvert \MC{T}_{k,T} \rvert)} \right)$. This has been proved in similar settings \citep{Abbasi-Yadkori2011, Agrawal2013,Vaswani2017,Kim2019}; for completeness, we present the proof.
Define $s_{i,k}(t) = \| b_i^c (t) \|_{B_j(k)^{-1}}$. Note that $s_{a(t), j_t} (t) = X_t$ and $ \sum_{t \in \MC{T}_{k,T}} \| X_{\tau} \|_{B_k(t)^{-1}} = $.
Then, $\| X_{\tau} \|_{B_k(t)^{-1}} = \sum_{t \in \MC{T}_{k,T}} s_{a(t), k} (t)$. Following the lines for equation 60 of \citet{Vaswani2017}, we can derive
\BEQ\label{eqn:detBjt}
\log \left[ \det( B_k(t+1) ) \right] \geq
 \log \left[  \det ( \lambda l_{kk} I_d ) \right] +
 \sum_{\tau \in \MC{T}_{k,t}} \log \left(1 + s_{a(\tau),k} (\tau)^2 \right).
\EEQ
On the other hand, the trace of $B_k(t+1)$ is
\BEQ\label{eqn:traceBjt}
\tr \left( B_k(t+1) \right) \leq 8\lvert \MC{T}_{k,T} \rvert + \lambda l_{kk} d,
\EEQ
where we used $\|X_{\tau} \| \leq 2$ by construction.
Plugging \eqref{eqn:detBjt} and \eqref{eqn:traceBjt} into the determinant-trace inequality $\left\{ \tr \left( B_k(t+1) \right) / d \right\}^d \geq \det \left( B_k(t+1) \right)$, equivalently $d \log \left\{ \tr \left( B_k(t+1) \right) / d \right\} \geq \log \det \left( B_k(t+1) \right)$, we obtain
\[
d \log \left( \frac{8\lvert \MC{T}_{k,T} \rvert}{d} + \lambda l_{kk} \right) \geq d \log (\lambda l_{kk}) + 
 \sum_{\tau \in \MC{T}_{k,t}} \log \left(1 + s_{a(\tau),k} (\tau)^2 \right),
\]
or,
\[
\sum_{\tau \in \MC{T}_{k,t}} \log \left(1 + s_{a(\tau),k} (\tau)^2 \right) \leq
 d \log \left( 1 + \frac{8\lvert \MC{T}_{k,T} \rvert}{d \lambda l_{kk}} \right).
\]
Now, we bound $\sum_{\tau \in \MC{T}_{k,t} }  s_{a(\tau),k} (\tau)^2$ by the result above. First, we have $s_{a(\tau),k} (\tau)^2 \in [0, 1/(\lambda l_{kk})]$ because
\[
s_{a(\tau),k} (\tau)^2 = b_{\tau}^T B_k(\tau)^{-1} b_{\tau}
\leq b_{\tau}^T \left( \lambda l_{kk} I_d \right)^{-1} b_{\tau}
\leq (\lambda l_{kk})^{-1}.
\]
Considering a function $f(t) = \log(1+t) / \left[ \lambda l_{kk} \log \left( 1 + (\lambda l_{kk})^{-1} \right) \right]$, $f$ satistfies $t \leq f(t)$ for all $t \in [0, 1/(\lambda l_{kk})]$. Therefore,
\BEA
\sum_{\tau \in \MC{T}_{k,t}} s_{a(\tau),k} (\tau)^2  &\leq&
 \frac{1}{\lambda l_{kk} \log \left( 1 + (\lambda l_{kk})^{-1} \right)} 
 \sum_{\tau \in \MC{T}_{k,t}} \log \left(1 + s_{a(\tau),k} (\tau)^2 \right) \NN \\
&\leq& \frac{d}{\lambda l_{kk} \log \left( 1 + (\lambda l_{kk})^{-1} \right)} 
 \log \left( 1 + \frac{\lvert \MC{T}_{k,T} \rvert}{d \lambda l_{kk}} \right). \NN
\EEA

Finally, from the Cauchy-Schwartz inequality and the result above,
\BEA
\sum_{\tau \in \MC{T}_{k,t}} s_{a(\tau),k} (\tau)  &\leq&
 \sqrt{8\lvert \MC{T}_{k,T} \rvert} \sqrt{ \sum_{\tau \in \MC{T}_{k,t}} s_{a(\tau),k} (\tau)^2 }  \NN \\
&\leq&
 \sqrt{ \frac{d\lvert \MC{T}_{k,T} \rvert}{\lambda l_{kk} \log \left( 1 + (\lambda l_{kk})^{-1} \right)} 
 \log \left( 1 + \frac{\lvert \MC{T}_{jk,t} \rvert}{d \lambda l_{kk}} \right) }. \NN
\EEA
Since $l_{kk}=1$ by the defintion of the random-walk Laplacian, 
\[
\sum_{\tau \in \MC{T}_{k,t}} s_{a(\tau),k} (\tau)  \leq
  \sqrt{ \frac{d\lvert \MC{T}_{k,T} \rvert}{\lambda  \log \left( 1 +  \frac{1}{\lambda} \right)} 
 \log \left( 1 + \frac{8\lvert \MC{T}_{k,T} \rvert}{d \lambda } \right) },
\]
which proves the claim and concludes the proof.
\end{proof}

\section{Regret bound when $\pi_i(t)$ is approximated by Monte Carlo sampling}\label{sec:regretMC}

In this section, we analyze the additional regret induced by  approximation and show that the regret upper bound of the alternative algorithm has the same order as the bound of {\sf SemiGraphTS}.

Our discussion is based on Algorithm \ref{algo:alternative}, a special case of  the {\sf SemiGraphTS} algorithm (Algorithm \ref{algo:SemiGraphTS}), that explicitly states that we use the Monte Carlo approximated values of $\pi_i(t)$ for action selection.
Before action selection, Algorithm \ref{algo:alternative} computes first the Monte Carlo approximates of $\pi_i(t)$. We denote the approximated value as $\WH{\pi}_i(t).$ Then, Algorithm \ref{algo:alternative} samples the arm from a multinomial distribution with size 1, say ${\rm Multinom}(\WH{\pi}_1(t), \cdots, \WH{\pi}_N(t))$.
In comparison, Algorithm \ref{algo:SemiGraphTS} samples
$a(t) \sim {\rm Multinom}(\pi_1(t), \cdots, \pi_N(t))$.

\begin{algorithm}[t]
\caption{A special case of  the SemiGraphTS algorithm that approximates $\pi_i(t)$ by the Monte Carlo sampling ({\sf SemiGraphTS-MC})} \label{algo:alternative}
\begin{algorithmic}[1] 
\State Fix $\lambda > 0$ and M. Set $B_j(1) = \lambda l_{jj} I_d$, $y_j(1) = 0_d$  and $v_j = (4R+12) \sqrt{  d  \log \left\{ ({24} T^4  / \delta) (1 + \lambda^{-1} )   \right\}  } + \sqrt{\lambda} (1 + \|\Delta_j\|)$ for $j=1,\ldots,n$.
\For{$t=1,2,\ldots, T$}
    \State{Observe $j_t$.}
    \For{$j=1,2,\ldots, n$}
        \If{$j \neq j_t$} 
            \State Update $B_j (t+1) \gets B_j (t)$, $\bar{\mu}_j (t+1) \gets \bar{\mu}_j (t)$, and $y_j (t+1) \gets y_j(t)$. 
        \Else
            \State $\WH{\mu}_j(t) \gets \bar{\mu}_j (t) - B_j (t)^{-1}   \sum_{k \neq j}  \lambda l_{jk} \bar{\mu}_{k} (t)$.
            \State $\Gamma_j(t) \gets B_j(t)  + \lambda^2 \! \sum_{k \neq j} \!l_{j k}^2 \! B_k(t)^{-1}$
            \For{$m=1,2,\cdots,M$}
            \State Sample $\TD{\mu}_j^m(t)$ from $\MC{N}_d( \WH{\mu}_j(t), v_j^2 \Gamma_j(t)^{-1} )$
            \EndFor
            \For{$i=1,2,\cdots,N$}
            \State Compute $\WH{\pi}_i(t)=\frac{1}{M}\sum_{m=1}^MI\left\{i=\argmax_k \{ b_k(t)^T \TD{\mu}_j^m(t) \}\right\}$
            \EndFor
            \State Sample $a(t)$ from ${\rm Multinom}(\WH{\pi}_1(t), \cdots, \WH{\pi}_N(t))$.
            \State $\bar{b}(t) \!\gets\! \sum_{i=1}^N\! \WH{\pi}_i (t) b_i (t)$ and $X_t \!\gets\!  b_{a(t)}(t) - \bar{b}(t)$.
            \State Update $B_j (t+1) \!\gets\!  B_j (t) \!+\! X_t  X_t^T \!+\! \sum_{i=1}^N \! \WH{\pi}_i(t) (b_i(t)-\bar{b}(t))(b_i(t)-\bar{b}(t))^T$, $y_j (t+1) \!\gets\! y_j (t) \!+\! 2 X_t r_{a(t), j} (t)$, and  $\bar{\mu}_j(t+1) \!\gets\! B_j (t+1)^{-1} y_j (t+1)$.
        \EndIf
    \EndFor
\EndFor
\end{algorithmic}
\end{algorithm}

We now discuss the regret bound for Algorithm \ref{algo:alternative}. We highlight the key differences from following the lines of Section \ref{sec:thm41}. Let the filtration $\MC{F}_{t-1}$ further include all Monte Carlo samples up to time $t-1$. 

For step (a), Theorem \ref{thm:Emuhat_t_semi} directly holds with $\pi_i(t)$'s replaced with $\WH{\pi}_i(t)$'s since the approximated values $\WH{\pi}_i(t)$'s are now the true probabilities of the arm selection.

Steps (b)-(e) in Section \ref{sec:thm41} exploited that the arm is selected form the exact probability. In other words, those results were derived if we select arm according to $\tilde{a}(t)=\argmax_{1 \leq i \leq N} \{ b_i(t)^T \TD{\mu}_{j_t}(t) \}$ (i.e., $\TD{a}(t) \sim {\rm Multinom}(\pi_1(t), \cdots, \pi_N(t))$). Now we show through an inductive argument that the remaining proofs are still valid with the new arm selection $a(t) \sim {\rm Multinom}(\WH{\pi}_1(t), \cdots, \WH{\pi}_N(t))$.

Suppose that until round $t-1$, we have sampled arms $a(\tau)\sim {\rm Multinom}(\WH{\pi}_1(\tau), \cdots, \WH{\pi}_N(\tau))$, $\tau=1,\cdots, t-1$. Then we have the desired high-probability upper bound for the estimate $\hat{\mu}_j(t)$ for every $j=1,\cdots,n$ (Theorem \ref{thm:Emuhat_t_semi}). Now suppose that at round $t$, we sample the arm $\tilde{a}(t)=\argmax_{1 \leq i \leq N} \{ b_i(t)^T \TD{\mu}_{j_t}(t) \}$. Then the proofs (b)-(e) go through, and by Lemma 13 we have,   
\begin{gather}
\MBE\left( (b_{a^*(t)}(t)^T\mu_{j_t}-b_{\tilde{a}(t)}(t)^T\mu_{j_t})I(E^{\hat{\mu}}(t)) \vert \MC{F}_{t-1} \right)
\NN \\
\leq 
\frac{5 g_{j_t}(T)}{p} \MBE\left( s^c_{{\tilde{a}}(t), {j_t}}(t) \vert \MC{F}_{t-1} \right)
+ \frac{4 g_{j_t}(T)}{pT^2}.
\end{gather}
Then, 
given  $\mathcal{F}_{t-1}$ such that $E^{\hat{\mu}}(t)$ is true,
\begin{align*}
& \MBE\left(regret(t)\vert\MC{F}_{t-1}\right) \\
&=\MBE\left( b_{a^*(t)}(t)^T\mu_{j_t}-b_{{a}(t)}(t)^T\mu_{j_t} \vert \MC{F}_{t-1} \right)\\
&=\MBE\left( (b_{a^*(t)}(t)^T\mu_{j_t}-b_{\tilde{a}(t)}(t)^T\mu_{j_t} )+ (b_{\tilde{a}(t)}(t)^T\mu_{j_t}-b_{{a}(t)}(t)^T\mu_{j_t}) \vert \MC{F}_{t-1} \right)\\
&\leq \frac{5 g_{j_t}(T)}{p} \MBE\left( s^c_{{\tilde{a}}(t), {j_t}}(t) \vert \MC{F}_{t-1} \right)
+ \frac{4 g_{j_t}(T)}{pT^2}+\MBE\left( b_{\tilde{a}(t)}(t)^T\mu_{j_t}-b_{{a}(t)}(t)^T\mu_{j_t} \vert \MC{F}_{t-1} \right)\\
&=\frac{5 g_{j_t}(T)}{p} \MBE\left( s^c_{{{a}}(t), {j_t}}(t) \vert \MC{F}_{t-1} \right)
+ \frac{4 g_{j_t}(T)}{pT^2}+\MBE\left( b_{\tilde{a}(t)}(t)^T\mu_{j_t}-b_{{a}(t)}(t)^T\mu_{j_t} \vert \MC{F}_{t-1} \right)\\
&~~~~ +\frac{5 g_{j_t}(T)}{p} \MBE\left( s^c_{{{\tilde{a}}}(t), {j_t}}(t)-s^c_{{{a}}(t), {j_t}}(t) \vert \MC{F}_{t-1} \right).
\end{align*}

As compared to Lemma 13 for Algorithm \ref{algo:SemiGraphTS}, we have two additional terms to bound; for step (f), those terms appear in the final  cumulative regret. We claim below that the cumulative sum of the two additional terms have lower order than the original regret bound of Algorithm \ref{algo:SemiGraphTS}. We first have,
\begin{align*}
& \MBE\left( b_{\tilde{a}(t)}(t)^T\mu_{j_t}-b_{{a}(t)}(t)^T\mu_{j_t} \vert \MC{F}_{t-1} \right) \\
&=\MBE\left( \sum_{i=1}^Nb_{i}(t)^T\mu_{j_t}I(\tilde{a}(t)=i)-\sum_{i=1}^Nb_{i}(t)^T\mu_{j_t}I(a(t)=i) \Big\vert \MC{F}_{t-1} \right)\\
&= \sum_{i=1}^Nb_{i}(t)^T\mu_{j_t}\pi_i(t)-\sum_{i=1}^Nb_{i}(t)^T\mu_{j_t}\MBE(\WH{\pi}_i(t) \vert \mathcal{F}_{t-1}) \\
&=\sum_{i=1}^Nb_{i}(t)^T\mu_{j_t}\pi_i(t)-\sum_{i=1}^Nb_{i}(t)^T\mu_{j_t}\pi_i(t)=0,
\end{align*}
which is due to unbiasedness of the Monte-Carlo estimate $\WH{\pi}_i(t)$. Since we also have  $||b_{\tilde{a}(t)}(t)^T\mu_{j_t}-b_{{a}(t)}(t)^T\mu_{j_t}|| \leq 2$, we can show from the Azuma-Hoeffiding inequality, with high probability,
\BEQ\label{eqn:stepF_final_additional_1}
\sum_{t=1}^T\left\{b_{\tilde{a}(t)}(t)^T\mu_{j_t}-b_{{a}(t)}(t)^T\mu_{j_t}\right\}\leq O(\sqrt{T}).
\EEQ
Similarly, we have
\begin{align*}
& \frac{5 g_{j_t}(T)}{p} \MBE\left( s^c_{{{\tilde{a}}}(t), {j_t}}(t)-s^c_{{{a}}(t), {j_t}}(t) \vert \MC{F}_{t-1} \right) \\
&= \frac{5 g_{j_t}(T)}{p} \MBE\left( \sum_{i=1}^Ns^c_{i, {j_t}}(t)I(\tilde{a}(t)=i)-\sum_{i=1}^Ns^c_{i, {j_t}}(t)I(a(t)=i) \Big\vert \MC{F}_{t-1} \right)\\
&=\frac{5 g_{j_t}(T)}{p}\sum_{i=1}^Ns^c_{i, {j_t}}(t)(\pi_i(t)-\pi_i(t))=0.
\end{align*}
Hence with high probability, 
\BEQ\label{eqn:stepF_final_additional_2}
\sum_{t=1}^T\frac{5 g_{j_t}(T)}{p}\left\{s^c_{{{\tilde{a}}}(t), {j_t}}(t)-s^c_{{{a}}(t), {j_t}}(t)\right\}\leq \frac{5 \max_j g_j(T)}{p}O(\sqrt{T}).
\EEQ
We remark that the right-hand sides of \eqref{eqn:stepF_final_additional_1} and \eqref{eqn:stepF_final_additional_2} does not depend on $d, n, N$ nor the graph structure. 
Therefore, our claim holds.

\end{appendices}




\end{document}